\pdfoutput=1
\documentclass[journal]{IEEEtran}
\ifCLASSINFOpdf
  \usepackage[pdftex]{graphicx}
   \usepackage{epstopdf}
\else
     \usepackage[dvips]{graphicx}
   \usepackage{epstopdf}
\fi
\usepackage[font=footnotesize]{subfig}

\usepackage{multirow}

\usepackage{rotating}

\begin{document}
%
\title{A Systematic Study of Online Class Imbalance Learning with Concept Drift}
%
%
%

\author{Shuo~Wang,~\IEEEmembership{Member,~IEEE,}
        Leandro L.~Minku,~\IEEEmembership{Member,~IEEE,}
        and Xin~Yao,~\IEEEmembership{Fellow,~IEEE}
\thanks{S. Wang and X. Yao are with the Centre of Excellence for Research in Computational Intelligence and Applications (CERCIA), School of Computer Science, The University of Birmingham, Edgbaston, Birmingham B15 2TT, UK. E-mail: \{S.Wang, X.Yao\}@cs.bham.ac.uk.}
\thanks{L. L. Minku is with Department of Informatics,
University of Leicester, Leicester LE1 7RH, UK. E-mail: leandro.minku@leicester.ac.uk.}}

%
%

\markboth{IEEE Transactions on Neural Networks and Learning Systems,~Vol.~xx, No.~x, February~2017}%
{Shell \MakeLowercase{\textit{et al.}}: Bare Demo of IEEEtran.cls for IEEE Journals}
%



\maketitle

\begin{abstract}
As an emerging research topic, online class imbalance learning often combines the challenges of both class imbalance and concept drift. It deals with data streams having very skewed class distributions, where concept drift may occur. It has recently received increased research attention; however, very little work addresses the combined problem where both class imbalance and concept drift coexist. As the first systematic study of handling concept drift in class-imbalanced data streams, this paper first provides a comprehensive review of current research progress in this field, including current research focuses and open challenges. Then, an in-depth experimental study is performed, with the goal of understanding how to best overcome concept drift in online learning with class imbalance. Based on the analysis, a general guideline is proposed for the development of an effective algorithm. 
\end{abstract}


%
\IEEEpeerreviewmaketitle

\section{Introduction}
\label{sec:intro}

With the wide application of machine learning algorithms to the real world, class imbalance and concept drift have become crucial learning issues. Applications in various domains such as risk management~\cite{Sousa2016pe}, anomaly detection~\cite{Meseguer2010bo}, software engineering~\cite{Wang2013hi}, and social media mining~\cite{Sun2016ba} are affected by both class imbalance and concept drift. Class imbalance happens when the data categories are not equally represented, i.e., at least one category is minority compared to other categories~\cite{He2009kx}. It can cause learning bias towards the majority class and poor generalization. Concept drift is a change in the underlying distribution of the problem, and is a significant issue specially when learning from data streams~\cite{Minku:2010uq}. It requires learners to be adaptive to dynamic changes. 

Class imbalance and concept drift can significantly hinder predictive performance, and the problem becomes particularly challenging when they occur simultaneously. This challenge arises from the fact that one problem can affect the treatment of the other. For example, drift detection algorithms based on the traditional classification error may be sensitive to the imbalanced degree and become less effective; and class imbalance techniques need to be adaptive to changing imbalance rates, otherwise the class receiving the preferential treatment may not be the correct minority class at the current moment.

Although there have been papers studying data streams with an imbalanced distribution and data streams with concept drift respectively, very little work discusses the cases when both class imbalance and concept drift exist. This paper aims to provide a systematic study of handling concept drift in class-imbalanced data streams. We focus on online (i.e. one-by-one) learning, which is a more difficult case than chunk-based learning, because only a single instance is available at a time. 

We first give a comprehensive review of current research progress in this field, including problem definitions, problem and approach categorization, performance evaluation and up-to-date approaches. It reveals new challenges and research gaps. Most existing work focuses on the concept drift in posterior probabilities (i.e. real concept drift~\cite{Gama2014re}, changes in $P\left(y \mid \mathbf{x} \right)$). The challenges in other types of concept drift have not been fully discussed and addressed. Especially, the change in prior probabilities $P\left( y \right)$ is closely related to class imbalance and has been overlooked by most existing work. Most proposed concept drift detection approaches are designed for and tested on balanced data streams. Very few approaches aim to tackle class imbalance and concept drift simultaneously. Among limited solutions, it is still unclear which approach is better and when. It is also unknown whether and how applying class imbalance techniques (e.g. resampling methods) affects concept drift detection and online prediction.

To fill in the research gaps, we then provide an experimental insight into how to best overcome concept drift in online learning with class imbalance, by focusing on three research questions: 1) what are the challenges in detecting each type of concept drift when the data stream is imbalanced? 2) Among the proposed methods designed for online class imbalance learning with concept drift, which one performs better for which type of concept drift? 3) Would applying class imbalance techniques (e.g. resampling methods) facilitate concept drift detection and online prediction? Six recent approaches, DDM-OCI~\cite{Wang2013bp}, LFR~\cite{Wang2015zo}, PAUC-PH~\cite{Brzezinski2015ib}~\cite{Brzezinski2017el}, OOB~\cite{Wang2014vb}, RLSACP~\cite{Ghazikhani2013wm} and ESOS-ELM~\cite{Mirza2015nt}, are compared and analyzed in depth under each of the three fundamental types of concept drift (i.e. changes in prior probability $P\left( y \right)$, class-conditional probability density function (pdf) $p\left(\mathbf{x}\mid y\right)$ and posterior probability $P\left(y \mid \mathbf{x} \right)$) in artificial data streams, as well as real-world data sets. To the best of our knowledge, they are the very few methods that are explicitly designed for online learning problems with class imbalance and concept drift so far.

Finally, based on the review and experimental results, we provide some guidelines for developing an effective algorithm for learning from imbalanced data streams with concept drift. We stress the importance of studying the mutual effect of class imbalance and concept drift. 

The contributions of this paper include: this is the first comprehensive study that looks into concept drift detection in class-imbalanced data streams; data problems are categorized in different types of concept drift and class imbalance with illustrative applications; existing approaches are compared and analysed systematically in each type; pros and cons of each approach are investigated; the results provide guidance for choosing the appropriate technique and developing better algorithms for future learning tasks; this is also the first work exploring the role of class imbalance techniques in concept drift detection, which sheds light on whether and how to tackle class imbalance and concept drift simultaneously.

The rest of this paper is organized as follows. Section~\ref{sec:framework} formulate the learning problem, including a learning framework and detailed problem descriptions and introduction of class imbalance and concept drift individually. Section~\ref{sec:overcomeboth} reviews the combined issue of class imbalance and concept drift, including example applications and existing solutions. Section~\ref{sec:exp} carries out the experimental study, aiming to find out the answers to the three research questions. Section~\ref{sec:con} draws the conclusions and points out potential future directions.

\section{Online Learning Framework with Class Imbalance and Concept Drift}
\label{sec:framework}

In data stream applications, data arrives over time in streams of examples or batches of examples. The information up to a specific time step $t$ is used to build/update predictive models, which then predict the new example(s) arriving at time step $t+1$. Learning under such conditions needs chunk-based learning or online learning algorithms, depending on the number of training examples available at each time step. According to the most agreed definitions~\cite{Minku:2010uq}~\cite{Ditzler2015hq}, chunk-based learning algorithms process a batch of data examples at each time step, such as the case of daily internet usage from a set of users; online learning algorithms process examples one by one and the predictive model is updated after receiving each example~\cite{Oza2001fk}, such as the case of sensor readings at every second in engineering systems. The term ``incremental learning" is also frequently used under this scenario. It is usually referred to as any algorithm that can process data streams with certain criteria met~\cite{Polikar2001oq}. 

On one hand, online learning can be viewed as a special case of chunk-based learning. Online learning algorithms can be used to deal with data coming in batches. They both build and continuously update a learning model to accommodate newly available data, and simultaneously maintain its performance on old data, giving rise to the stability-plasticity dilemma~\cite{Grossberg1988as}. On the other hand, the way of designing online and chunk-based learning algorithms can be very different~\cite{Minku:2010uq}. Most chunk-based learning algorithms are not suitable for online learning tasks, because batch learners process a chunk of data each time, possibly using an offline learning algorithm for each chunk. Online learning requires the model being adapted immediately upon seeing the new example, and the example is then immediately discarded, which allows to process high-speed data streams.  From this point of view, designing online learning algorithm can be more challenging but so far has received much less attention than the other.

First, the online learner needs to learn from a single data example, so it needs a more sophisticated training mechanism. Second, data streams are often non-stationary (concept drift). The limited availability of training examples at the current moment in online learning hinders the detection of such changes and the application of techniques to overcome the change. Third, it is often seen that data is class imbalanced in many classification tasks, such as the fault detection task in an engineering system, where the fault is always the minority. Class imbalance aggravates the learning difficulty~\cite{He2009kx} and complicates the data status~\cite{Wang2013po}. However, there is a severe lack of research addressing the combined issue of class imbalance and concept drift in online learning. 

To fill in this research gap, this paper aims at a comprehensive review of the work done to overcome class imbalance and concept drift, a systematic study of learning challenges, and an in-depth analysis of the performance of current approaches. We begin by formalizing the learning problem in this section. 

\subsection{Learning Procedure}
\label{subsec:procedure}

In supervised online classification, suppose a data generating process provides a sequence of examples $\left( \mathbf{x_t}, y_t \right)$ arriving one at a time from an unknown probability distribution $p_t\left(x,y \right)$. $\mathbf{x_t}$ is the input vector belonging to an input space $X$, and $y_t$ is the corresponding class label belonging to the label set $Y = \left\{ c_1, \ldots, c_N \right\}$. We build an online classifier $F$ that receives the new input $\mathbf{x_t}$ at time step $t$ and then makes a prediction. The predicted class label is denoted by $\hat{y}_t$. After some time, the classifier receives the true label $y_t$, used to evaluate the predictive performance and further train the classifier. This whole process will be repeated at following time steps. It is worth pointing out that we do not assume new training examples always arrive at regular and pre-defined intervals here. In other words, the actual time interval between time step $t$ and $t+1$ may be different from the actual time interval between $t+1$ and $t+2$.

One challenge arises when data is class imbalanced. Class imbalance is an important data feature, commonly seen in applications such as spam filtering~\cite{Nishida2008fk} and fault diagnosis~\cite{Meseguer2010bo}~\cite{Wang2013hi}. It is the phenomenon when some classes of data are highly under-represented (i.e. minority) compared to other classes (i.e. majority). For example, if $P\left ( c_i \right )\ll P\left ( c_j \right )$, then $c_j$ is a majority class and $c_i$ is a minority class. The difficulty in learning from imbalanced data is that the relatively or absolutely underrepresented class cannot draw equal attention to the learning algorithm, which often leads to very specific classification rules or missing rules for this class without much generalization ability for future prediction. It has been well-studied in offline learning~\cite{Japkowicz:2002ul}, and has attracted growing attention in data stream learning in recent years~\cite{Hoens2012uq}. 

In many applications, such as energy forecasting and climate data analysis~\cite{Monteiro2009nw}, the data generator operates in nonstationary environments. It gives rise to another challenge, called ``concept drift". It means that the probability density function (pdf) of the data generating process is changing over time. For such cases, the fundamental assumption of traditional data mining -- the training and testing data are sampled from the same static and unknown distribution -- does not hold anymore. Therefore, it is crucial to monitor the underlying changes, and adapt the model to accommodate the changes accordingly. 

\begin{figure}[htp]
\centerline{
\includegraphics[width=3in]{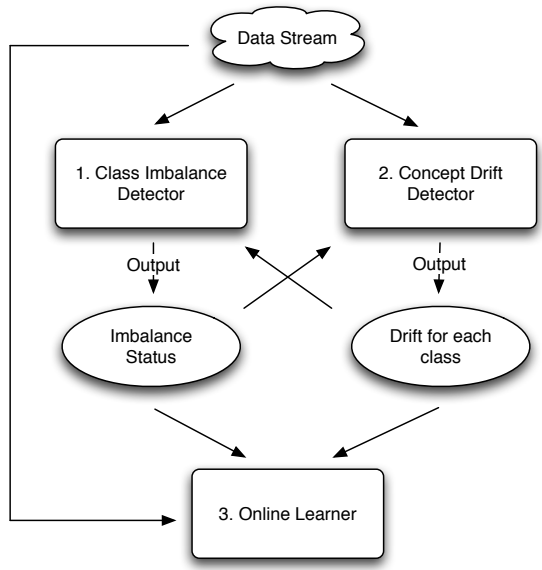}}
\caption{Learning framework for online class imbalance learning~\cite{Wang2013po}.}
\label{fig:framework}
\end{figure}

When both issues exist, the online learner needs to be carefully designed for effectiveness, efficiency and adaptivity. An online class imbalance learning framework was proposed in~\cite{Wang2013po} as a guide for algorithm design. The framework breaks down the learning procedure into three modules -- a class imbalance detector, a concept drift detector and an adaptive online learner, as illustrated in Fig.~\ref{fig:framework}. 

The class imbalance detector reports the current class imbalance status of data streams. The concept drift detector captures concept drifts involving changes in classification boundaries. Based on the information provided by the first two modules, the adaptive online learner determines when and how to respond to the detected class imbalance and concept drift, in order to maintain its performance. The learning objective of an online class imbalance algorithm can be described as ``recognizing minority-class data effectively, adaptively and timely without sacrificing the performance on the majority class"~\cite{Wang2013po}.

\subsection{Problem Descriptions}
\label{subsec:probdesp}

A more detailed introduction about class imbalance and concept drift is given here individually, including the terminology, research focuses and state-of-the-art approaches. The purpose of this section is to understand the fundamental issues that we need to take extra care of in online class imbalance learning. We also aim at understanding whether and how the current research in class imbalance learning and concept drift detection are individually related to their combined issue elaborated later in Section~\ref{sec:overcomeboth}, rather than to provide an exhaustive list of approaches in the literature. Among others, we will answer the following questions: \textit{can existing class imbalance techniques process data streams? Would existing concept drift detectors be able to handle imbalanced data streams?}

\subsubsection{\textbf{Class imbalance}}

In class imbalance problems, the minority class is usually much more difficult or expensive to be collected than the majority class, such as the spam class in spam filtering and the fraud class in credit card application. Thus, misclassifying a minority-class example is more costly. Unfortunately, the performance of most conventional machine learning algorithms is significantly compromised by class imbalance, because they assume or expect balanced class distributions or equal misclassification costs. Their training procedure with the aim of maximizing overall accuracy often leads to a high probability of the induced classifier predicting an example as the majority class, and a low recognition rate on the minority class. In reality, it is common to see that the majority class has accuracy close to 100\% and the minority class has very low accuracy between 0\%-10\%~\cite{Kubat:1998bs}. The negative effect of class imbalance on classifiers, such as decision trees~\cite{Japkowicz:2002ul}, neural networks~\cite{Visa:2005ai}, k-Nearest Neighbour (kNN)~\cite{Kubat:1997yg}~\cite{Batista:2004oq}~\cite{Zhang:2003ve} and SVM~\cite{Yan:2003vn}~\cite{Wu:2003qe}, has been studied. A classifier that provides a balanced degree of predictive performance for all classes is required. The major research questions in this area are summarized and answered as follows:\\

\noindent
(a) \textit{How do we define the imbalanced degree of data?}

It seems to be a trivial question. However, there is no consensus on the definition in the literature. To describe how imbalanced the data is, researchers choose to use the percentage of the minority class in the data set~\cite{Hulse:2007eu}, the size ratio between classes~\cite{Lopez2013ls}, or simply a list of the number of examples in each class~\cite{Chawla:2002yq}. The coefficient of variance is used in~\cite{Hoens2012lq}, which is less straightforward. The description of imbalance status may not be a crucial issue in offline learning, but becomes more important in online learning, because there is no static data set in online scenarios. It is necessary to have some measurement automatically describing the up-to-date imbalanced degree and techniques monitoring the changes in class imbalance status. This will help the online learner to decide when and how to tackle class imbalance. The issue of changes in class imbalance status is relevant to concept drift, which will be further discussed in the next subsection. 

To define the imbalanced degree suitable for online learning, a real-time indicator was proposed -- time-decayed class size~\cite{Wang2013po}, expressing the size percentage of each class in the data stream. It is updated incrementally at each time step by using a time decay (forgetting) factor, which emphasizes the current status of data and weakens the effect of old data. Based on this, a class imbalance detector was proposed to determine which classes should be regarded as the minority/majority and how imbalanced the current data stream is, and then used for designing better online classifiers~\cite{Wang2014vb}~\cite{Wang2013hi}. The merit of this indicator is that it is suitable for data with arbitrary number of classes. 

\noindent
(b) \textit{When does class imbalance matter?}

It has been shown that class imbalance is not the only problem responsible for the performance reduction of classifiers. Classifiers' sensitivity to class imbalance also depends on the complexity and overall size of the data set. Data complexity comprises issues such as overlapping~\cite{Batista:2005uq}~\cite{Prati:2004kx} and small disjuncts~\cite{Jo2004bh}. The degree of overlapping between classes and how the minority class examples distribute in data space aggravate the negative effect of class imbalance. The small disjunct problem is associated with the within-class imbalance~\cite{Japkowicz:2003pd}. Regarding the size of the training data, a very large domain has a good chance that the minority class is represented by a reasonable number of examples, and thus may be less affected by imbalance than a small domain containing very few minority class examples. In other words, the rarity of the minority class can be in a relative or absolute sense in terms of the number of available examples~\cite{He2009kx}.

In particular, authors in~\cite{Napierala2012bo}~\cite{Napierala2016gr} distinguished and analysed four types of data distributions in the minority class -- safe, borderline, outliers and rare examples. Safe examples are located in the homogenous regions populated by the examples from one class only; borderline examples are scattered in the boundary regions between classes, where the examples from both classes overlap; rare examples and outliers are singular examples located deeper in the regions dominated by the majority class. Borderline, rare and outlier data sets were found to be the real source of difficulties in learning imbalanced data sets offline, which have also been shown to be the harder cases in online applications~\cite{Wang2014vb}. Therefore, for any developed algorithms dealing with imbalanced data online, it is worth discussing their performance on data with different types of distributions. 

\noindent
(c) \textit{How can we tackle class imbalance effectively (state-of-the-art solutions)?}

A number of algorithms have been proposed to tackle class imbalance at the data and algorithm levels. Data-level algorithms include a variety of resampling techniques, manipulating training data to rectify the skewed class distributions. They oversample minority-class examples (i.e. expanding the minority class), undersample majority-class examples (i.e. shrinking the majority class), or combine both, until the data set is relatively balanced. Random oversampling and random undersampling are the simplest and most popular resampling techniques, where examples are randomly chosen to be added or removed. There are also smart resampling techniques (a.k.a guided resampling). For example, SMOTE~\cite{Chawla:2002yq} is a widely used oversampling method, which generates new minority-class data points based on the similarities between original minority-class examples in the feature space. Other smart oversampling techniques include Borderline-SMOTE~\cite{Han:2005sf}, ADASYN~\cite{He:2008cr}, MWMOTE~\cite{Barua2014mq}, to name but a few. Smart undersampling techniques include Tomek links~\cite{TOMEK:1976la}, One-sided selection~\cite{Kubat:1997kx}, Neighbourhood cleaning rule~\cite{Jorma:2001kx}, etc. The effectiveness of resampling techniques have been proved in real-world applications~\cite{Hao2014po}. They work independently of classifiers, and are thus more versatile than algorithm-level methods. The key is to choose an appropriate sampling rate~\cite{Estabrooks:2004ve}, which is relatively easy for two-class data sets, but becomes more complicated for multi-class data sets~\cite{Saez2016yq}. Empirical studies have been carried out to compare different resampling methods~\cite{Hulse:2007eu}. Particularly, it is shown that smart resampling techniques are not necessarily superior to random oversampling and undersampling; besides, they cannot be applied to online scenarios directly, because they work on a static data set for the relation among the training examples. Some initial effort has been made recently, to extend smart resampling techniques to online learning~\cite{Mao2015xp}.

Algorithm-level methods address class imbalance by modifying their training mechanism with the direct goal of better accuracy on the minority class, including one-class learning~\cite{Japkowicz:1995qf}, cost-sensitive learning~\cite{Liu:2006yq} and threshold methods~\cite{Weiss:2003eu}. They require different treatments for specific kinds of learning algorithms. In other words, they are algorithm-dependent, so they are not as widely used as data-level methods. Some online cost-sensitive methods have been proposed, such as CSOGD~\cite{Wang2014fn} and RLSACP~\cite{Ghazikhani2013wm}. They are restricted to the perceptron-based classifiers, and require pre-defined misclassification costs of classes that may or may not be updated during the online learning. 

Finally, ensemble learning (also known as multiple classifier systems)~\cite{Polikar:2006xy} has become a major category of approaches to handling class imbalance~\cite{Galar:2011uq}. It combines multiple classifiers as base learners and aims to outperform every one of them. It can be easily adapted for emphasizing the minority class by integrating different resampling techniques~\cite{Li:2007hc}~\cite{Liu:2009kx}~\cite{Chawla:2003yq}~\cite{Blaszczynski2015ge}  or by making base classifiers cost-sensitive~\cite{Joshi:2001kx}~\cite{Chawla:2007kx}~\cite{Guo:2004qe}~\cite{Fan:1999rm}. A few ensemble methods are available for online class imbalance learning, such as OOB and UOB~\cite{Wang2014vb} applying random oversampling and undersampling in Online Bagging~\cite{Oza:2005ve}, and WOS-ELM~\cite{Mirza2013la} training a set of cost-sensitive online extreme learning machines. 

It is worth pointing out that, the aforementioned online learning algorithms designed for imbalanced data are not suitable for non-stationary data streams. They do not involve any mechanism handling drifts that affect classification boundaries, although OOB and UOB can detect and react to class imbalance changes. 

\noindent
(d) \textit{How do we evaluate the performance of class imbalance learning algorithms?}

Traditionally, overall accuracy and error rate are the most frequently used metrics of performance evaluation. However, they are strongly biased towards the majority class when data is imbalanced. Therefore, other performance measures have been adopted. Most studies concentrate on two-class problems. By convention, the minority class is treated to be the positive, and the majority class is treated to be the negative. Table~\ref{tab:confusion} illustrates the confusion matrix of a two-class problem, producing four numbers on testing data. 

\begin{table}[htp]
\caption{Confusion matrix for a two-class problem.}
\label{tab:confusion}
\centering
\begin{tabular}{|c|c|c|}
\hline
 & Predicted as positive & Predicted as negative\\ \hline
Actual positive & True positive (TP) & False negative (FN) \\ 
Actual negative & False positive (FP) & True negative (TN) \\ \hline
\end{tabular}
\end{table}

From the confusion matrix, we can derive the expressions for \textit{recall} and \textit{precision}:

\begin{equation}
\label{eq:recall}
recall = \frac{TP}{TP+FN},
\end{equation}
\begin{equation}
\label{eq:precision}
precision = \frac{TP}{TP+FP}.
\end{equation}

Recall (i.e. TP rate) is a measure of completeness -- the proportion of positive class examples that are classified correctly to all positive class examples. Precision is a measure of exactness -- the proportion of positive class examples that are classified correctly to the examples predicted as positive by the classifier. The learning objective of class imbalance learning is to improve recall without hurting precision. However, improving recall and precision can be conflicting. Thus, F-measure is defined to show the trade-off between them. 

\begin{equation}
\label{eq:F}
Fm = \frac{\left ( 1+\beta^2 \right )\cdot recall\cdot precision}{\beta^2\cdot precision+recall},
\end{equation}

where $\beta$ corresponds to the relative importance of recall and precision. It is usually set to 1. Kubat et al.~\cite{Kubat:1997kx} proposed to use G-mean to replace overall accuracy: 

\begin{equation}
\label{eq:G}
Gm = \sqrt{\frac{TP}{TP+FN}\times \frac{TN}{TN+FP}}.
\end{equation}

It is the geometric mean of positive accuracy (i.e. TP rate) and negative accuracy (i.e. TN rate). A good classifier should have high accuracies on both classes, and thus a high G-mean.

According to~\cite{He2009kx}, any metric that uses values from both rows of the confusion matrix for addition (or subtraction) will be inherently sensitive to class imbalance. In other words, the performance measure will change as class distribution changes, even though the underlying performance of the classifier does not. This performance inconsistency can cause problems when we compare different algorithms over different data sets. Precision and F-measure, unfortunately, are sensitive to the class distribution. Therefore, recall and G-mean are better options. 

To compare classifiers over a range of sample distributions, AUC (abbr. of the Area Under the ROC curve) is the best choice. A ROC curve depicts all possible trade-offs between TP rate and FP rate, where FP rate = $FP/\left( TN+FP \right)$. TP rate and FP rate can be understood as the benefits and costs of classification with respect to data distributions. Each point on the curve corresponds to a single trade-off. A better classifier should produce a ROC curve closer to the top left corner. AUC represents a ROC curve as a single scalar value by estimating the area under the curve, varying in [0, 1]. It is insensitive to the class distribution, because both TP rate and FP rate use values from only one row of the confusion matrix. AUC is usually generated by varying the classification decision threshold for separating positive and negative classes in the testing data set~\cite{Maloof:2003dq}~\cite{Fawcett:2006uq}. In other words, calculating AUC requires a set of confusion matrices. Therefore, unlike other measures based on a single confusion matrix, AUC cannot be used as an evaluation metric in online learning without memorizing data. Although a recent study has modified AUC for evaluating online classifiers~\cite{Brzezinski2015ib}, it still needs to collect recently received examples. 

The properties of the above measures are summarized in Table~\ref{tab:imbalance-metric}. They are defined under the two-class context. They cannot be used to evaluate multi-class data directly, except for recall. Their multi-class versions have been developed~\cite{Sokolova2009dl}~\cite{Sun:2006pd}~\cite{Hand:2001kx}. The ``multi-class" and ``online" columns in the table show whether the corresponding measure can be used directly without modification in multi-class and online data scenarios.  

\begin{table}[htp]
\caption{Performance evaluation measures for class imbalance problems.}
\label{tab:imbalance-metric}
\centering
\begin{tabular}{|c|c|c|c|}
\hline
Measures& Multi-class & Online & Sensitive to\\ 
&&&Imbalance\\ \hline
recall & yes & yes & no\\ \hline
precision & no~\cite{Sokolova2009dl} & yes & yes\\ \hline
Fm & no~\cite{Sokolova2009dl} & yes & yes\\ \hline
Gm & yes~\cite{Sun:2006pd} & yes & no\\ \hline
AUC & no (See MAUC~\cite{Hand:2001kx}) & no (See PAUC~\cite{Brzezinski2015ib}) & no\\ \hline
\end{tabular}
\end{table}

\subsubsection{\textbf{Concept drift}}

Concept drift is said to occur when the joint probability $P\left( \mathbf{x},y \right)$ changes~\cite{Gama2014re}~\cite{Minku2012vn}~\cite{Minku2010uq2}. The key research topics in this area include:\\

\noindent
(a) \textit{How many types of concept drift are there? Which type is more challenging?}

\begin{figure}[htp]
\centerline{
\subfloat[Original Distribution]{
\includegraphics[width=1.8in]{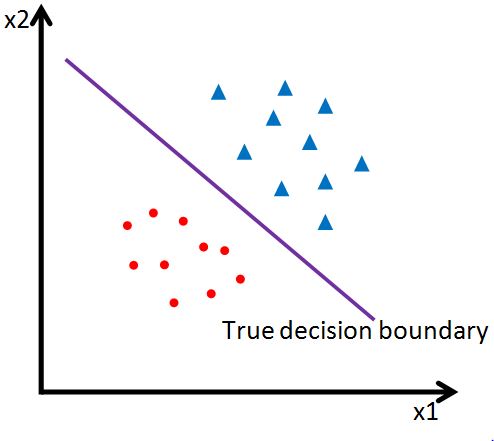}}
\subfloat[$P\left( y \right)$ drift]{
\includegraphics[width=1.8in]{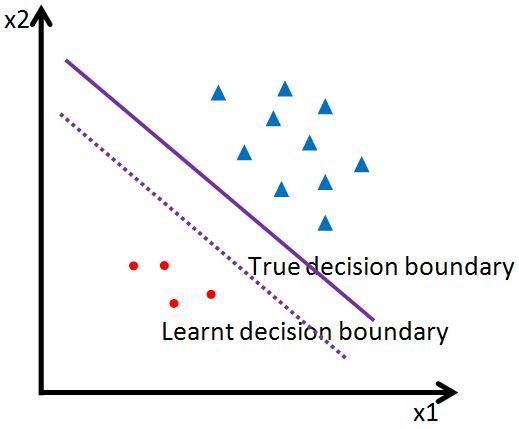}}}

\centerline{
\subfloat[$p\left( \mathbf{x}\mid y \right)$ drift]{
\includegraphics[width=1.8in]{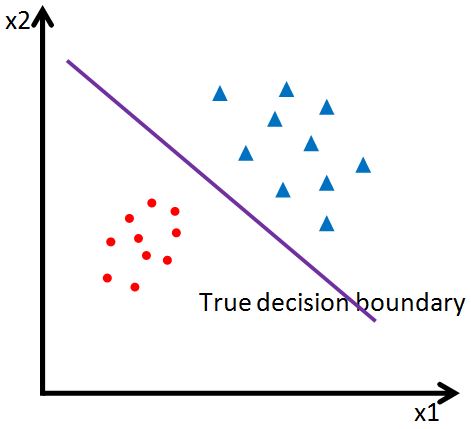}}
\subfloat[$P\left( y\mid \mathbf{x} \right)$ drift]{
\includegraphics[width=1.8in]{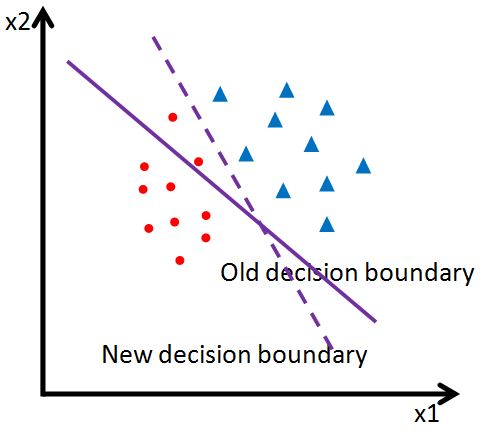}}}
\caption{Illustration of 3 concept drift types.}
\label{fig:drift}
\end{figure}

Concept drift can manifest three fundamental forms of changes corresponding to the three major variables in the Bayes' theorem~\cite{Kelly1999fs}: 1) a change in prior probability $P\left( y \right)$; 2) a change in class-conditional pdf $p\left(\mathbf{x}\mid y \right)$; 3) a change in posterior probability $P\left( y\mid \mathbf{x} \right)$. The three types of concept drift are illustrated in Figure~\ref{fig:drift}. Comparing to the original data distribution shown in Figure~\ref{fig:drift}(a),

Fig.~\ref{fig:drift}(b) shows the $P\left( y \right)$ type of concept drift without affecting $p\left(  \mathbf{x}\mid y \right)$ and $P\left( y\mid  \mathbf{x} \right)$. The decision boundary remains unaffected. The prior probability of the circle class is reduced in this example. Such change can lead to class imbalance. A well-learnt discrimination function may drift away from the true decision boundary, due to the imbalanced class distribution. 

Fig.~\ref{fig:drift}(c) shows the $p\left(  \mathbf{x}\mid y \right)$ type of concept drift without affecting $P\left( y \right)$ and $P\left( y\mid  \mathbf{x} \right)$. The true decision boundary remains unaffected. Elwell and Polikar claimed that this type of drift is the result of an incomplete representation of the true distribution in current data, which simply requires providing supplemental data information to the learning model~\cite{Elwell2011dq}.
 
Fig.~\ref{fig:drift}(d) shows the $P\left( y\mid  \mathbf{x} \right)$ type of concept drift. The true boundary between classes changes after the drift, so that the previously learnt discrimination function does not apply any more. In other words, the old function becomes unsuitable or partially unsuitable, and the learning model needs to be adapted to the new knowledge. 

The posterior distribution change clearly indicates the most fundamental change in the data generating function. This is classified as \textit{real concept drift}. The other two types belong to \textit{virtual concept drift}~\cite{Hoens2012uq}, which does not change the decision (class) boundaries. In practice, one type of concept drift may appear in combination with other types. 

Existing studies primarily focus on the development of drift detection methods and techniques to overcome the real drift. There is a significant lack of research on virtual drift, which can also deteriorate classification performance. As illustrated in Fig.~\ref{fig:drift}(b), even though these types of drift do not affect the true decision boundaries, they can cause a well-learnt decision boundary to become unsuitable. Unfortunately, the current techniques for handling real drift may not be suitable for virtual drift, because they present very different learning difficulties and require different solutions. For instance, the methods for handling real drift often choose to reset and retrain the classifier, in order to forget the old concept and better learn the new concept. This is not an appropriate strategy for data with virtual drift, because the examples from previous time steps may still remain valid and help the current classification in virtual drift cases. It would be more effective and efficient to calibrate the existing classifier than retraining it. Besides, techniques for handling real drift typically rely on feedback about the performance of the classifier, while techniques for handling virtual drift can operate without such feedback~\cite{Gama2014re}. From our point of view, all three types are equally important. Particularly, the two virtual types require more research effort than currently dedicated work by our community. A systematic study of the challenges in each type will be given in Section~\ref{sec:exp}.

\begin{table*}[htp]
\caption{Categorization of concept drift techniques. See~\cite{Ditzler2015hq} for the full list of techniques under each category.}
\label{tab:driftdetector}
\centering
\begin{tabular}{|c|c|l|}
\hline
\multirow{14}{*}{\textbf{Active}} & \multirow{6}{*}{\textbf{Step1. Change}} & \textbf{Hypothesis tests}: assess the validity of a hypothesis by comparing the distributions of two sets of fix-length \\ 
&& data sequences. \\ \cline{3-3}
&& \textbf{Change-point methods}: identify the change point by analyzing all possible partitions of a fixed data sequence. \\ \cline{3-3}
&\multirow{2}{*}{\textbf{detection}}& \textbf{Sequential hypothesis tests}: provide a one-off detection of change or no change, by inspecting incoming \\
&& examples one by one (sequentially). \\ \cline{3-3}
&& \textbf{Change detection tests}: analyze the statistical behavior of streams of data in a fully sequential manner, such \\
&& as a feature value or classification error. They are either based on a pre-defined threshold or some statistical \\
&& features representing current data. \\ \cline{2-3}

& \multirow{5}{*}{\textbf{Step2. Classifier}} & \textbf{Windowing}: the classifier is retrained based on a window with up-to-date examples. The window length can \\
&& be either fixed or adaptive. \\ \cline{3-3}
&& \textbf{Weighting}: all received examples are weighted according to time or classification error, which are then used to \\
&\multirow{1}{*}{\textbf{adaptation}}& update the classifier. \\ \cline{3-3}
&& \textbf{Random Sampling}: the examples used to retrain the classifier are randomly chosen based on certain rules. \\ \cline{3-3}
&& \textbf{Ensemble}: build a new model in the classifier for the new concept.  \\ \hline

\multirow{2}{*}{\textbf{Passive}} & \multicolumn{2}{|l|}{\textbf{Single classifier}: update a single classifier, such as decision trees, online information network, and extreme learning machine.} \\ \cline{2-3}
& \multicolumn{2}{|l|}{\textbf{Ensemble}: add, remove or modify the models in an ensemble classifier.} \\ \hline
\end{tabular}
\end{table*}

Concept drift has further been characterized by its speed, severity, cyclical nature, etc. A detailed and mutually exclusive categorization can be found in~\cite{Minku2010uq2}. For example, according to speed, concept drift can be either abrupt, when the generating function is changed suddenly (usually within one time step), or gradual, when the distribution evolves slowly over time. They are the most commonly discussed types in the literature, because the effectiveness of drift detection methods can vary with the drifting speed. While most methods are quite successful in detecting abrupt drifts, as future data is no longer related to old data~\cite{Ditzler2013nh}, gradual drifts are often more difficult, because the slow change can delay or hide the hint left by the drift. We can see some drift detection methods specifically designed for gradual concept drift, such as Early Drift Detection method (EDDM)~\cite{Baena-Garca2006tg}.

\noindent
(b) \textit{How can we tackle concept drift effectively (state-of-the-art solutions)?}

There is a wide range of algorithms for learning in non-stationary environments. Most of them assume and specialize in some specific types of concept drift, although real-world data often contains multiple types. They are commonly categorized into two major groups: active vs. passive approaches, depending on whether an explicit drift detection mechanism is employed. Active approaches (also known as trigger-based approaches) determine whether and when a drift has occurred before taking any actions. They operate based on two mechanisms -- a change detector aiming to sense the drift accurately and timely, and an adaptation mechanism aiming to maintain the performance of the classifier by reacting to the detected drift. Passive approaches (also known as adaptive classifiers) evolve the classifier continuously without an explicit trigger reporting the drift. A comprehensive review of up-to-date techniques tackling concept drift is given by Ditzler et al.~\cite{Ditzler2015hq}. They further organise these techniques based on their core mechanisms, summarized in Table~\ref{tab:driftdetector}. This table will help us to understand how online class imbalance algorithms are designed, which will be introduced in details in Section~\ref{sec:overcomeboth}. There exist other ways to classify the proposed algorithms, such as Gama et al.'s taxonomy based on the four modules of an adaptive learning system~\cite{Gama2014re}, and Webb et al.'s quantitative characterization~\cite{Webb2016jq}. This paper adopts the one proposed by Ditzler et al.~\cite{Ditzler2015hq} for its simplicity. 

The best algorithm varies with the intended applications. A general observation is that, while active approaches are quite effective in detecting abrupt drift, passive approaches are very good at overcoming gradual drift~\cite{Elwell2011dq}~\cite{Ditzler2015hq}. It is worth noting that most algorithms do not consider class imbalance. It is unclear whether they will remain effective if data becomes imbalanced. For example, some algorithms determine concept drift based on the change in the classification error, including OLIN~\cite{Cohen2008vn}, DDM~\cite{Gama2004kl} and PERM~\cite{Harel2014oa}. As we have explained in Section~\ref{subsec:probdesp} 1), the classification error is sensitive to the imbalance degree of data, and does not reflect the performance of the classifier very well when there is class imbalance. Therefore, these algorithms may not perform well when concept drift and class imbalance occur simultaneously. Some other algorithms are specifically designed for data streams coming in batches, such as AUE~\cite{Brzezinski2014fo} and the Learn++ family~\cite{Elwell2011dq}. These algorithms cannot be applied to online cases directly. 

\noindent
(c) \textit{How do we evaluate the performance of concept drift detectors and online classifiers?}

To fully test the performance of drift detection approaches (especially an active detector), it is necessary to discuss both data with artificial concept drifts and real-world data with unknown drifts. Using data with artificial concept drifts allows us to easily manipulate the type and timing of concept drifts, so as to obtain an in-depth understanding of the performance of approaches under various conditions. Testing on data from real-world problems helps us to understand their effectiveness from the practical point of view, but the information about when and how concept drift occurs is unknown in most cases. The following aspects are usually considered to assess the accuracy of active drift detectors. Their measurement is based on data with artificial concept drifts where drifts are known.  

\begin{itemize}
\item True detection rate: the possibility of detecting the true concept drift. It shows the accuracy of the detection approach. 
\item False alarm rate: the possibility of reporting a concept drift that does not exist (false-positive rate). It characterizes the costs and reliability of the detection approach. 
\item Delay of detection: an estimate of how many time steps are required on average to detect a drift after the actual occurrence. It reflects how much time would be taken before the drift is detected. 
\end{itemize}

Wang and Abraham~\cite{Wang2015zo} use a histogram to visualize the distribution of detection points from the drift detection approach over multiple runs. It reflects all the three aspects above in one plot. It is worth nothing that there are trade-offs between these measures. For example, an approach with a high true detection rate may produce a high false alarm rate. A very recent algorithm, Hierarchical Change-Detection Tests (HCDTs), was proposed to explicitly deal with the trade-off~\cite{Alippi2017hw}.

After the performance of drift detection approaches is better understood, we need to quantify the effect of those detections on the performance of predictive models. All the performance metrics introduced in the previous section of ``class imbalance" can be used. The key question here is how to calculate them in the streaming settings with evolving data. The performance of the classifier may get better or worse every now and then. There are two common ways to depict such performance over time -- holdout and prequential evaluation~\cite{Gama2014re}. 

Holdout evaluation is mostly used when the testing data set (holdout set) is available in advance. At each time step or every few time steps, the performance measures are calculated based on the valid testing set, which must represent the same data concept as the training data at that moment. However, this is a very rigorous requirement for data from real-world applications.

In prequential evaluation, data received at each time step is used for testing before it is use for training. From this, the performance measures can be incrementally updated for evaluation and comparison. This strategy does not require a holdout set, and the model is always tested on unseen data. 

When the data stream is stationary, the prequential performance measures can be computed based on the accumulated sum of a loss function from the beginning of the training. However, if the data stream is evolving, the accumulated measure can mask the fluctuation in performance and the adaptation ability of the classifier. For example, consider that an online classifier correctly predicts 90 out of 100 examples received so far (90\% accuracy on data with the original concept). Then, an abrupt concept drift occurs at time step 101, which makes the classifier only correctly predict 3 out of 10 examples from the new concept (30\% accuracy on data with the new concept). If we use the accumulated measure based on all the historical data, the overall accuracy will be 93/110, which seems to be high but does not reflect the true performance on the new data concept. This problem can be solved by using a sliding window or a time-based fading factor that weigh observations~\cite{Gama2013qp}.

\section{Overcoming Class Imbalance and Concept Drift Simultaneously}
\label{sec:overcomeboth}

Following the review of class imbalance and concept drift in Section~\ref{sec:framework}, this section reviews the combined issue, including example applications and existing solutions. When both exist, one problem affects the treatment of the other. For example, the drift detection algorithms based on the traditional classification error may be sensitive to imbalanced degree and become less effective; the class imbalance techniques need to be adaptive to changing $P\left( y \right)$, otherwise the class receiving the preferential treatment may not be the correct minority class at the current moment. Therefore, their mutual effect should be considered during the algorithm design.

\subsection{Illustrative Applications}
\label{subsec:application}

The combined problems of concept drift and class imbalance have been found in many real-world applications. Three examples are given here, to help us understand each type of concept drift. 

\subsubsection{Environment monitoring with $P\left( y \right)$ drift}

Environment monitoring systems usually consist of various sensors generating streaming data in high speed. Real-time prediction is required. For example, a smart building has sensors deployed to monitor hazardous events. Any sensor fault can cause catastrophic failures. Machine learning algorithms can be used to build models based on the sensor information, aiming to predict faults in sensors accurately and timely~\cite{Wang2013hi}. First, the data is characterized by class imbalance, because obtaining a fault in such systems can be very expensive. Examples representing faults are the minority. Second, the number of faults varies with the faulty condition. If the damage gets worse over time, the faults will occur more and more frequently. It implies a prior probability change, a type of virtual concept drift.  

\subsubsection{Spam filtering with $p\left( \mathbf{x}\mid y \right)$ drift}

Spam filtering is a typical classification problem involving class imbalance and concept drift~\cite{Lindstrom2010qg}. First of all, the spam class is the minority and suffers from a higher misclassification cost. Second, the spammers are actively working on how to break through the filter. It means that the adversary actions are adaptive. For example, one of the spamming behaviours is to change email content and presentation in disguise, implying a possible class-conditional pdf ($p\left( \mathbf{x}\mid y \right)$) change~\cite{Gama2014re}. 

\subsubsection{Social media analysis with $P\left( y\mid \mathbf{x} \right)$ drift}

Social media (e.g. twitter, facebook) is becoming a valuable source of timely information on the internet. It attracts a growing number of people, sharing, communicating, connecting and creating user-generated data. Consider the example where a company would like to make relevant product recommendations to people who have shown some type of interest in their tweets. Machine learning algorithms can be used to discover who is interested in the product from the large amount of tweets~\cite{Li2012la}. The number of users who have shown the interest is always very small. Their information tends to be overwhelmed by other unrelated messages. Thus, it is utterly important to overcome the imbalanced distribution and discover the hidden information. Another challenge is users' interest changing from time to time. Users may lose their interest in the current trendy product very quickly, causing posterior probability ($P\left( y\mid \mathbf{x} \right)$) changes.

Although the above examples are associated with only one type of concept drift, different types often coexist in real-world problems, which are hard to know in advance. For the example of spam filtering, which email belongs to spam also depends on users' interpretation. Users may re-label a particular category of normal emails as spam, which indicates a posterior probability change. 

\subsection{Approaches to Tackling Both Class Imbalance and Concept Drift}
\label{subsec:approach}

Some research efforts have been made to address the joint problem of concept drift and class imbalance, due to the rising need from practical problems~\cite{Pan2015wp}~\cite{Sousa2016pe}. Uncorrelated Bagging is one of the earliest algorithms, which builds an ensemble of classifiers trained on a more balanced set of data through resampling and overcomes concept drift passively by weighing the base classifier based on their discriminative power~\cite{Gao2008uq}~\cite{Gao2007fk}~\cite{Wu2014nq}. Selectively recursive approaches SERA~\cite{Chen2009dz} and REA~\cite{Chen2011tg} use similar ideas to Uncorrelated Bagging of building an ensemble of weighted classifiers, but with a ``smarter" oversampling technique. Learn++.CDS and Learn++.NIE are more recent algorithms, which tackle class imbalance through the oversampling technique SMOTE~\cite{Chawla:2002yq} or a sub-ensemble technique, and overcome concept drift through a dynamic weighting strategy~\cite{Ditzler2013mk}. HUWRS.IP~\cite{Hoens2013gh} improves HUWRS~\cite{Hoens:2011ys} to deal with imbalanced data streams by introducing an instance propagation scheme based on a Na\"{i}ve Bayes classifier, and uses Hellinger distance as a weighting measure for concept drift detection. This method relies on finding examples that are similar to the current minority-class concept, which however may not exist. So, Hellinger Distance Decision Tree (HDDT) was proposed to use Hellinger distance as the decision tree splitting criteria that is imbalance-insensitive~\cite{Pozzolo2014wl}. All these approaches belong to chunk-based learning algorithms. Their core techniques work when a batch of data is received at each time step, i.e. they are not suitable for online processing. Developing a true online algorithm for concept drift is very challenging because of the difficulties in measuring minority-class statistics using only one example at a time~\cite{Ditzler2015hq}. 

To handle class imbalance and concept drift in an online fashion, a few methods have been proposed recently. Drift Detection Method for Online Class Imbalance (DDM-OCI)~\cite{Wang2013bp} is one of the very first algorithms detecting concept drift actively in imbalanced data streams online. It monitors the reduction in minority-class recall (i.e. true positive rate). If there is a significant drop, a drift will be reported. It was shown to be effective in cases when minority-class recall is affected by the concept drift, but not when the majority class is mainly affected. A Linear Four Rates (LFR) approach was then proposed to improve DDM-OCI, which monitors four rates from the confusion matrix -- minority-class recall and precision and majority-class recall and precision, with statistically-supported bounds for drift detection~\cite{Wang2015zo}. If any of the four rates exceeds the bound, a drift will be confirmed. Instead of tracking several performance rates for each class, prequential AUC (PAUC)~\cite{Brzezinski2015ib}~\cite{Brzezinski2017el} was proposed as an overall performance measure for online scenarios, and was used as the concept drift indicator in Page-Hinkley (PH) test~\cite{Page1954qg}. However, it needs access to historical data. DDM-OCI, LFR and PAUC-based PH test are active drift detectors designed for imbalanced data streams, and are independent of classification algorithms. They aim at concept drift with classification boundary changes by default. Therefore, if a concept drift is reported, they will reset and retrain the online model. Although these drift detectors are designed for imbalanced data, they themselves do not handle class imbalance. It is still unclear how they perform when working with class imbalance techniques.  

\begin{table*}[htp]
\caption{Online approaches to tackling concept drift and class imbalance, and their properties.}
\label{tab:onlinemethod}
\centering
\begin{tabular}{|c|c|c|c|c|c|c|}
\hline
Approaches & Category? & Class  & Access to & Additional data? & Multi-class? & $P\left( y \right)$ drift? \\ 
& & imbalance? & old data? & & & \\ \hline
DDM-OCI~\cite{Wang2013bp} & Active (change detection test + windowing) & No & No & No & No & No \\ \hline
LFR~\cite{Wang2015zo} & Active (change detection test + windowing) & No & No & No & No & No \\ \hline
PAUC-PH~\cite{Brzezinski2015ib} & Active (change detection test + windowing) & No & Yes & No & No & No \\ \hline
RLSACP~\cite{Ghazikhani2013wm}/ONN~\cite{Ghazikhani2014qb} & Passive (single classifier) & Yes & Yes & No & No & Yes \\ \hline
ESOS-ELM~\cite{Mirza2015nt} & Passive+Active (ensemble) & Yes & No & Yes & No & No \\ \hline
OOB/UOB using CID~\cite{Wang2014vb} & Active (weighting) & Yes & No & No & No & Yes \\ \hline
\end{tabular}
\end{table*}

Besides the above active approaches, the perceptron-based algorithms RLSACP~\cite{Ghazikhani2013wm}, ONN~\cite{Ghazikhani2014qb} and ESOS-ELM~\cite{Mirza2015nt} adapt the classification model to non-stationary environments passively, and involve mechanisms to overcome class imbalance. RLSACP and ONN are single-model approaches with the same general idea. Their error function for updating the perceptron weights is modified, including a forgetting function for model adaptation and an error weighting strategy as the class imbalance treatment. The forgetting function has a pre-defined form, allowing the old data concept to be forgotten gradually. The error weights in RLSACP are incrementally updated based either on the classification performance or the imbalance rate from recently received data. It was shown that weight updating based on the imbalance rate leads to better performance.  

ESOS-ELM is an ensemble approach, maintaining a set of online sequential extreme learning machines (OS-ELM)~\cite{Liang2006bp}. For tackling class imbalance, resampling is applied in a way that each OS-ELM is trained with approximately equal number of minority- and majority-class examples. For tackling concept drift, voting weights of base classifiers are updated according to their performance G-mean on a separate validation data set from the same environment as the current training data. In addition to the passive drift detection technique, ESOS-ELM includes an independent module -- ELM-store, to handle recurring concept drift. ELM-store maintains a pool of weighted extreme learning machines (WELM)~\cite{Mirza2013la} to retain old information. It adopts a threshold-based technique and hypothesis testing to detect abrupt and gradual concept drift actively. If a concept drift is reported, a new WELM will be built and kept in ELM-store. If any stored model performs better than the current OS-ELM ensemble, indicating a possible recurring concept, it will be introduced in the ensemble. ESOS-ELM assumes the imbalance rate is known in advance and fixed. It needs a separate data set for initializing OS-ELMs and WELMs, which must include examples from all classes. It is also necessary to have validation data sets reflecting every data concept for concept drift detection, which can be a quite restrictive requirement for real-world data. 

With a different goal of concept drift detection from the above, a class imbalance detection (CID) approach was proposed, aiming at $P\left( y \right)$ changes~\cite{Wang2013po}. It reports the current imbalance status and provides information of which classes belong to the minority and which classes belong to the majority. Particularly, a key indicator is the real-time class size $w_k^{(t)}$, the percentage of class $c_k$ at time step t. When a new example $\mathbf{x_t}$ arrives, $w_k^{(t)}$ is incrementally updated by the following equation~\cite{Wang2013po}:

\begin{equation}
w_k^{(t)}=\theta w_k^{(t-1)}+\left ( 1-\theta \right ) \left[ \left (\mathbf{x_t},c_k \right ) \right], (k=1,\ldots,N)
\label{eq:wk}
\end{equation}

where $\left[ \left( \mathbf{x_t}, c_k \right) \right] = 1$ if the true class label of $\mathbf{x_t}$ is $c_k$, and 0 otherwise. $\theta$ $\left(0<\theta<1\right)$ is a pre-defined time decay (forgetting) factor, which reduces the contribution of older data to the calculation of class sizes along with time. It is independent of learning algorithms, so it can be used with any type of online classifiers. For example, it has been used in OOB and UOB~\cite{Wang2014vb} for deciding the resampling rate adaptively and overcoming class imbalance effectively over time. OOB and UOB integrate oversampling and undersampling respectively into ensemble algorithm Online Bagging (OB)~\cite{Oza:2005ve}. Oversampling and undersampling are one of the simplest and most effective techniques of tackling class imbalance~\cite{Hulse:2007eu}.

The properties of the above online approaches are summarized in Table~\ref{tab:onlinemethod}, answering the following six questions in order: 
\begin{itemize}
\item How do they handle concept drift (the type based on the categorization in Table~\ref{tab:driftdetector})?
\item Do they involve any class imbalance technique to improve the predictive performance of online models, in addition to concept drift detection?
\item Do they need access to previously received data?
\item Do they need additional data sets for initialisation or validation?
\item Can they handle data streams with more than two classes (multi-class data)?
\item Do they involve any mechanism handling $P\left( y \right)$ drift?
\end{itemize}

\section{Performance Analysis}
\label{sec:exp}

With a complete review of online class imbalance learning, we aim at a deep understanding of concept drift detection in imbalanced data streams and the performance of existing approaches introduced in Section~\ref{subsec:approach}. Three research questions will be looked into through experimental analysis: \textit{1) what are the difficulties in detecting each type of concept drift?} Little work has given separate discussions on the three fundamental types of concept drift, especially the $P\left( y \right)$ drift. It is important to understand their differences, so that the most suitable approaches can be used for the best performance. \textit{2) Among existing approaches designed for imbalanced data streams with concept drift, which approach is better and when?} Although a few approaches have been proposed for the purpose of overcoming concept drift and class imbalance, it is still unclear how well they perform for each type of concept drift. \textit{3) Whether and how do class imbalance techniques affect concept drift detection and online prediction?} No study has looked into the mutual effect of applying class imbalance techniques and concept drift detection methods. Understanding the role of class imbalance techniques will help us to develop more effective concept drift detection methods for imbalanced data. 

\subsection{Data Sets}
\label{subsec:data}

For an accurate analysis and comparable results, we choose two most commonly used artificial data generators, SINE1~\cite{Gama2004kl} and SEA~\cite{Street2001bh}, to produce imbalanced data streams containing three simulated types of concept drift. This is one of the very few studies that individually discuss $P\left( y \right)$, $p\left( \mathbf{x}\mid y \right)$ and $P\left(y \mid \mathbf{x} \right)$ types of concept drift in depth. In addition, each generator produces two data streams with a different drifting speed -- abrupt and gradual drifts. The drifting speed is defined as the inverse of the time taken for a new concept to completely replace the old one~\cite{Minku2010uq2}. According to speed, drifts can be either abrupt, when the generating function is changed completely in only one time step, or gradual, otherwise. The data streams with a gradual concept drift are denoted by `g' in the following experiment, i.e. SINE1g~\cite{Baena-Garca2006tg} and SEAg. Every data stream has 3000 time steps, with one concept drift starting at time step 1501. The new concept in SINE1 and SEA fully takes over the data stream from time step 1501; the concept drift in SINE1g and SEAg takes 500 time steps to complete, which means that the new concept fully replaces the old one from time step 2001. The detailed settings for generating each type of concept drift are included in the individual subsections. 

After the detailed analysis of the three types of concept drift, three real-world data sets are included in our experiment with unknown concept drift, which are PAKDD 2009 credit card data (PAKDD)~\cite{Linhart2010ln}, Weather data~\cite{Ditzler2013nh} and UDI TweeterCrawl data~\cite{LiWDWC12}. Data in PAKDD are collected from the private label credit card operation of a Brazilian retail chain. The task of this problem is to identify whether the client has a good or bad credit. The ``bad" credit is the minority class, taking 19.75\% of the provided modelling data. Because the data have been collected from a time interval in the past, gradual market change occurs. The Weather data set aims to predict whether rain precipitation was observed on each day, with inherent seasonal changes. The class of ``rain" is the minority at IR of 31\%. The original Tweet data include 50 million tweets posted mainly from 2008 to 2011. The task is to predict the tweet topic. We choose a time interval, containing 8774 examples and covering seven tweet topics~\cite{Wang2016wl}. Then, we further reduce it to 2-class data by using only two out of seven topics for our experiment. These real-world data will help us to understand the effectiveness of existing concept drift and class imbalance approaches in practical scenarios, which usually have more complex data distributions and concept drift. 

\subsection{Experimental and Evaluation Settings}
\label{subsec:setting}

\begin{table*}[htp]
\caption{Artificial data streams with $P\left( y \right)$ concept drift.}
\label{tab:pydata}
\centering
\begin{tabular}{|c|c|c|c|c|c|c|c|c|}
\hline
ID & Data& Speed & \multicolumn{3}{|c|}{Class +1} & \multicolumn{3}{|c|}{Class -1}\\ \cline{4-9}
&&& Concept & Old $P\left( y \right)$ & New $P\left( y \right)$ & Concept & Old $P\left( y \right)$ & New $P\left( y \right)$ \\ \hline
1&SINE1 & Abrupt & \multirow{2}{*}{Points below $y=\sin \left ( x \right )$} & \multirow{2}{*}{0.1} & \multirow{2}{*}{0.9} & \multirow{2}{*}{Points above or on $y=\sin \left ( x \right )$} & \multirow{2}{*}{0.9} & \multirow{2}{*}{0.1} \\ \cline{1-3}
2&SINE1g & Gradual &&&&&& \\ \hline
3&SEA & Abrupt & \multirow{2}{*}{$x_1+x_2 \leq 7$} & \multirow{2}{*}{0.5} & \multirow{2}{*}{0.1} & \multirow{2}{*}{$x_1+x_2 > 7$} & \multirow{2}{*}{0.5} & \multirow{2}{*}{0.9} \\ \cline{1-3}
4&SEAg & Gradual &&&&&& \\ \hline
\end{tabular}
\end{table*}

The approaches listed in Table~\ref{tab:onlinemethod}, which are explicitly designed for the combined problem of class imbalance and concept drift, are discussed in our experiment. For the three active drift detection methods -- DDM-OCI, LFR and PAUC-PH, they are used with the traditional Online Bagging (abbr. OB)~\cite{Oza:2005ve} and OOB with CID~\cite{Wang2014vb} respectively for classification. Because OOB applies oversampling to overcome class imbalance and OB does not, it can help us to observe the role of class imbalance techniques (oversampling in our experiment) in concept drift detection. UOB is not chosen, for the consideration that undersampling may cause unstable performance which may indirectly affect our observation~\cite{Wang2014vb}. Between RLSACP and ONN, due to their similarity and the more theoretical support in RLSACP, only RLSACP is included in our experiment. 

Considering RLSACP and ESOS-ELM are perceptron-based methods, we use the Multilayer Perceptron (MLP) classifier as the base learner of OB and OOB. The number of neurons in the hidden layer of MLPs is set to the average of the number of attributes and classes in data, which is also the number of perceptrons in RLSACP and ESOS-ELM. All ensemble methods maintain 15 base learners. For ESOS-ELM, we disable the ``ELM-Store", which is designed for recurring concept drift; we allow that its ensemble size can grow to 20. In addition, ESOS-ELM requires an initialisation data set to initialize ELMs, and validation data sets to adjust misclassification costs. When dealing with artificial data, we use the first 100 examples to initialize ESOS-ELM, and generate a separate validation data set for each concept stage. We track the performance of all the methods from time step 101.

In summary, ten algorithms join the comparison from Table~\ref{tab:onlinemethod}: OB, OOB, DDM-OCI+OB/OOB, PAUC-PH+OB/OOB, LFR+OB/OOB, RLSACP and ESOS-ELM. OB is the baseline without involving any class imbalance and concept drift techniques. 

To evaluate the effectiveness of concept drift detection methods and online learners, we adopt prequential test (as described in Section~\ref{sec:framework}) for its simplicity and popularity. Prequential recall of each class (defined in Eq.~\ref{eq:recall}) and prequential G-mean (defined in Eq.~\ref{eq:G}) are tracked over time for comparison, because they are insensitive to imbalance rates. When discussing the generated artificial data sets with ground truth known, we also compare the true detection rate (abbr. TDR), total number of false alarms (abbr. FA) and delay of detection (abbr. DoD) (as defined in Section~\ref{sec:framework}) among methods using any of the three active drift detectors (i.e. DDM-OCI, LFR and PAUC-PH). The calculation of TDR, FA and DoD is based on the following understanding: before a real concept drift occurs, all the reported alarms are considered as false alarms; after a real concept drift occurs, the first detection is seen as the true alarm; after that and before the next new real concept drift, the consequent detections are considered as false alarms. 

Furthermore, because we are particularly interested in how the learner performs on the new data concept in the artificial data sets, we calculate the average recall and G-mean over all the time steps before the concept drift starts and after the concept drift completely ends. It is worth noting that the recall and G-mean values are reset to 0 when the drift starts and ends for an accurate analysis. We use the Wilcoxon Sign Rank test at the confidence level of 95\% as our significance test in this paper. 

\subsection{Comparative Study on Artificial Data}
\label{subsec:artificial_analysis}

\noindent C.1. $\mathbf P\left( y \right)$ \textbf{Concept Drift}

This section focuses on the $P\left(  y \right)$ type of concept drift, without $p\left( \mathbf{x} \mid y \right)$ and $P\left(  y \mid \mathbf{x} \right)$ changes. Data streams SINE1 and SINE1g have a severe class imbalance change, in which the minority (majority) class during the first half of data streams becomes the majority (minority) during the latter half. SEA and SEAg have a less severe change, in which the data stream presented to be balanced during the first half becomes imbalanced during the latter half. The concrete setting for each data stream is summarized in Table~\ref{tab:pydata}. 

Table~\ref{tab:pyDetectors} compares the detection performance of the three active concept drift detectors, in terms of TDR, FA and DoD. The first column is the data ID number, as denoted in Table~\ref{tab:pydata}. We can see that DDM-OCI and LFR are sensitive to class imbalance changes in data. They present very high true detection rate; especially, LFR has 100\% TDR in all cases regardless of whether resampling is used to tackle class imbalance. PAUC-PH does not report any concept drift, showing 0\% TDR in all cases. This is because DDM-OCI and LFR use time-decayed metrics as the indicator of concept drift, which have higher sensitivity to performance change in general than the prequential AUC used by PAUC-PH. LFR shows even higher TDR than DDM-OCI, because it tracks four rates in the confusion matrix instead of one. For the same reason, DDM-OCI and LFR have a higher chance of issuing false alarms than PAUC-PH. For DDM-OCI, oversampling in OOB increases the probability of reporting a concept drift by observing TDR in SEA and SEAg, compared to OB. This is because more examples are used for training in OOB, which improves the performance on the minority class for concept drift detection. 

\begin{table}[htp]
\caption{Performance of the 3 active concept drift detectors on artificial data with $P\left( y \right)$ changes: TDR, FA and DoD. The `-' symbol indicates that no concept drift is detected.}
\label{tab:pyDetectors}
\centering
\begin{tabular}{|c|c|c|c|c|}
\hline
 & Method & TDR & FA & DoD \\ \hline
 \multirow{6}{*}{\begin{turn}{90}SINE1\end{turn}} & DDM-OCI+OB & 100\% & 0 & 94 \\ \cline{2-5}
&DDM-OCI+OOB & 100\% & 2.22 & 45 \\ \cline{2-5}
&LFR+OB & 100\% & 24 & 91\\ \cline{2-5}
&LFR+OOB & 100\% & 26.16 & 63\\ \cline{2-5}
&PAUC-PH+OB & 0\% & 1.03 & -\\ \cline{2-5}
&PAUC-PH+OOB & 0\% & 1.28 & - \\ \hline

 \multirow{6}{*}{\begin{turn}{90}SINE1g\end{turn}} & DDM-OCI+OB & 100\% & 1.09 & 281 \\ \cline{2-5}
&DDM-OCI+OOB & 100\% & 4.38 & 118 \\ \cline{2-5}
&LFR+OB & 100\% & 18.01 & 383 \\ \cline{2-5}
&LFR+OOB & 100\% & 21.15 & 153 \\ \cline{2-5}
&PAUC-PH+OB & 0\% & 1 & - \\ \cline{2-5}
&PAUC-PH+OOB & 0\% & 1 & - \\ \hline

\multirow{6}{*}{\begin{turn}{90}SEA\end{turn}} &DDM-OCI+OB & 45\% & 11.9 & 255\\ \cline{2-5}
&DDM-OCI+OOB & 94\% & 14.1 & 301\\ \cline{2-5}
&LFR+OB & 100\% & 0.73 & 35\\ \cline{2-5}
&LFR+OOB & 100\% & 6.51 & 45\\ \cline{2-5}
&PAUC-PH+OB & 0\% & 1 & -\\ \cline{2-5}
&PAUC-PH+OOB & 0\% & 1 & - \\ \hline

\multirow{6}{*}{\begin{turn}{90}SEAg\end{turn}} & DDM-OCI+OB & 92\% & 15.1 & 80\\ \cline{2-5}
 & DDM-OCI+OOB & 100\% & 16.56 & 93\\ \cline{2-5}
 & LFR+OB & 100\% & 2.27 & 121\\ \cline{2-5}
 & LFR+OOB & 100\% & 6.3 & 324\\ \cline{2-5}
 & PAUC-PH+OB & 0\% & 1 & -\\ \cline{2-5}
 & PAUC-PH+OOB & 0\% & 1.01 & - \\ \hline
\end{tabular}
\end{table}

Table~\ref{tab:pyLearners} compares recall and G-mean of all models over the new data concept, i.e. performance over time steps 1501-3000 for data streams with an abrupt change and performance over time steps 2001-3000 for data streams with a gradual change, showing whether and how well the drift detector can help with learning after concept drift is completed. The first column is the data ID number, as denoted in Table~\ref{tab:pydata}. In SINE1 and SINE1g, the negative class presents to be the minority after the change; in SEA and SEAg, the positive class presents to be the minority after the change. 

\begin{table}[htp]
\caption{Performance of online learners on artificial data with $P\left( y \right)$ changes: means and standard deviations of average recall of each class and average G-mean over the new data concept. The significantly best values among all methods are shown in bold italics.}
\label{tab:pyLearners}
\centering
\resizebox{\columnwidth}{!}{
\begin{tabular}{|c|c|c|c|c|}
\hline
 & Method & Class+1 Recall & Class-1 Recall & G-mean \\ \hline
 \multirow{10}{*}{\begin{turn}{90}SINE1\end{turn}} & DDM-OCI+OB & 0.887$\pm$0.004 & 0.170$\pm$0.009 & 0.317$\pm$0.009 \\ \cline{2-5}
&DDM-OCI+OOB & 0.979$\pm$0.007 & 0.049$\pm$0.016 & 0.188$\pm$0.033 \\ \cline{2-5}
&LFR+OB & 0.870$\pm$0.004 & 0.183$\pm$0.019 & 0.334$\pm$0.022\\ \cline{2-5}
&LFR+OOB & 0.952$\pm$0.011 & 0.061$\pm$0.023 & 0.221$\pm$0.042\\ \cline{2-5}
&PAUC-PH+OB & 0.889$\pm$0.004 & 0.168$\pm$0.008 & 0.316$\pm$0.007\\ \cline{2-5}
&PAUC-PH+OOB & \textbf{0.992$\pm$0.002} & 0.692$\pm$0.013 & 0.828$\pm$0.008 \\ \cline{2-5}
&RLSACP & 0.962$\pm$0.004 & 0.072$\pm$0.014 & 0.217$\pm$0.026 \\ \cline{2-5}
&ESOS-ELM & 0.176$\pm$0.136 & \textbf{0.999$\pm$0.001} & 0.358$\pm$0.192 \\ \cline{2-5}
&OB & 0.889$\pm$0.004 & 0.170$\pm$0.009 & 0.318$\pm$0.009 \\ \cline{2-5}
&OOB & \textbf{0.992$\pm$0.002} & 0.699$\pm$0.014 & \textbf{0.832$\pm$0.008}\\ \hline

 \multirow{10}{*}{\begin{turn}{90}SINE1g\end{turn}} & DDM-OCI+OB & \textbf{1.000$\pm$0.000} & 0.000$\pm$0.000 & 0.000$\pm$0.000\\ \cline{2-5}
&DDM-OCI+OOB & 0.997$\pm$0.004 & 0.008$\pm$0.005 & 0.050$\pm$0.016\\ \cline{2-5}
&LFR+OB & 0.972$\pm$0.006 & 0.031$\pm$0.027 & 0.138$\pm$0.079\\ \cline{2-5}
&LFR+OOB & 0.956$\pm$0.011 & 0.036$\pm$0.026 & 0.150$\pm$0.076\\ \cline{2-5}
&PAUC-PH+OB & \textbf{1.000$\pm$0.000} & 0.000$\pm$0.000 & 0.000$\pm$0.000\\ \cline{2-5}
&PAUC-PH+OOB & 0.989$\pm$0.001 & 0.708$\pm$0.002 & \textbf{0.835$\pm$0.002}\\ \cline{2-5}
&RLSACP & \textbf{1.000$\pm$0.000} & 0.000$\pm$0.001 & 0.002$\pm$0.013\\ \cline{2-5}
&ESOS-ELM & 0.109$\pm$0.102 & \textbf{0.997$\pm$0.000} & 0.273$\pm$0.165\\ \cline{2-5}
&OB & \textbf{1.000$\pm$0.000} & 0.000$\pm$0.000 & 0.000$\pm$0.000\\ \cline{2-5}
&OOB & 0.989$\pm$0.002 & 0.709$\pm$0.002 & \textbf{0.835$\pm$0.001} \\ \hline

\multirow{10}{*}{\begin{turn}{90}SEA\end{turn}} &DDM-OCI+OB & 0.003$\pm$0.031 & \textbf{0.999$\pm$0.000} & 0.007$\pm$0.055\\ \cline{2-5}
&DDM-OCI+OOB & 0.146$\pm$0.072 & 0.965$\pm$0.013 & 0.344$\pm$0.086\\ \cline{2-5}
&LFR+OB & 0.020$\pm$0.009 & 0.996$\pm$0.001 & 0.113$\pm$0.053\\ \cline{2-5}
&LFR+OOB & 0.059$\pm$0.031 & 0.981$\pm$0.007 & 0.221$\pm$0.054\\ \cline{2-5}
&PAUC-PH+OB & 0.323$\pm$0.010 & 0.995$\pm$0.001 & 0.559$\pm$0.009\\ \cline{2-5}
&PAUC-PH+OOB & 0.514$\pm$0.015 & 0.943$\pm$0.007 & \textbf{0.688$\pm$0.010}\\ \cline{2-5}
&RLSACP & 0.021$\pm$0.023 & 0.993$\pm$0.007 & 0.070$\pm$0.077\\ \cline{2-5}
&ESOS-ELM & \textbf{0.608$\pm$0.214} & 0.829$\pm$0.140 & \textbf{0.681$\pm$0.142}\\ \cline{2-5}
&OB & 0.324$\pm$0.009 & 0.996$\pm$0.001 & 0.561$\pm$0.008\\ \cline{2-5}
&OOB & 0.515$\pm$0.016 & 0.945$\pm$0.006 & \textbf{0.689$\pm$0.010} \\ \hline

\multirow{10}{*}{\begin{turn}{90}SEAg\end{turn}} & DDM-OCI+OB & 0.040$\pm$0.073 & 0.998$\pm$0.001 & 0.124$\pm$0.136\\ \cline{2-5}
&DDM-OCI+OOB & 0.142$\pm$0.071 & 0.973$\pm$0.014 & 0.334$\pm$0.096\\ \cline{2-5}
&LFR+OB & 0.003$\pm$0.006 & \textbf{0.999$\pm$0.000} & 0.019$\pm$0.035\\ \cline{2-5}
&LFR+OOB & 0.076$\pm$0.084 & 0.976$\pm$0.018 & 0.217$\pm$0.123\\ \cline{2-5}
&PAUC-PH+OB & 0.365$\pm$0.029 & 0.997$\pm$0.000 & 0.600$\pm$0.023\\ \cline{2-5}
&PAUC-PH+OOB & 0.489$\pm$0.024 & 0.951$\pm$0.011 & \textbf{0.679$\pm$0.017}\\ \cline{2-5}
&RLSACP & 0.002$\pm$0.006 & \textbf{0.999$\pm$0.001} & 0.011$\pm$0.035\\ \cline{2-5}
&ESOS-ELM & \textbf{0.562$\pm$0.208} & 0.809$\pm$0.143 & 0.646$\pm$0.130\\ \cline{2-5}
&OB & 0.371$\pm$0.029 & 0.997$\pm$0.001 & 0.605$\pm$0.023\\ \cline{2-5}
&OOB & 0.484$\pm$0.032 & 0.951$\pm$0.012 & \textbf{0.675$\pm$0.022} \\ \hline
\end{tabular}
}
\end{table}

\begin{table*}[htp]
\caption{Artificial data streams with $p\left( \mathbf{x} \mid y \right)$ concept drift.}
\label{tab:pxydata}
\centering
\begin{tabular}{|c|c|c|c|c|c|c|}
\hline
ID & Data& Speed & \multicolumn{2}{|c|}{Class +1} & \multicolumn{2}{|c|}{Class -1}\\ \cline{4-7}
&&& Old concept & New concept & Old concept & New concept \\ \hline
1&SINE1 & Abrupt & \multirow{2}{*}{Points below $y=\sin \left ( x \right )$} & \multirow{2}{*}{Points below $y=\sin \left ( x \right )$} & Points above or on $y=\sin \left ( x \right )$ & Points above or on $y=\sin \left ( x \right )$ \\ \cline{1-3}
2&SINE1g & Gradual &&&and $P\left ( x<0.5 \right ) = 0.9$&and $P\left ( x<0.5 \right ) = 0.1$ \\ \hline
3&SEA & Abrupt & \multirow{2}{*}{$x_1+x_2 \leq 7$} & \multirow{2}{*}{$x_1+x_2 \leq 7$} & $x_1+x_2 > 7$ & $x_1+x_2 > 7$ \\ \cline{1-3}
4&SEAg & Gradual &&&and $P\left ( x_1<5 \right ) = 0.9$&and $P\left ( x_1<5 \right ) = 0.1$ \\ \hline
\end{tabular}
\end{table*}

In terms of minority-class recall, we can see that ESOS-ELM performs the significantly best, but ESOS-ELM sacrifices majority-class recall, especially in SINE1 and SINE1g. In terms of G-mean, OOB and OOB using PAUC-PH perform the significantly best, which shows they can best balance the performance between classes. It is worth noting that PAUC-PH is the drift detection method with 0\% TDR based on Table~\ref{tab:pyDetectors}. It means that OOB plays the main role in learning. It also explains that OOB and OOB using PAUC-PH have very close performance. All the OB and OOB models using the other active drift detectors do not show competitive recall and G-mean. Especially for those using DDM-OCI and LFR, the high number of false alarms causes too much resetting and performance loss; OOB can increase the chance of producing a false alarm, because more minority-class examples join the training. 

Therefore, we conclude that, for $P\left( y \right)$ type of concept drift, it is not necessary to apply any drift detection techniques that are not specifically designed for class imbalance changes; the use of these drift detectors could be even detrimental to the predictive performance due to false alarms and performance resetting; the adaptive resampling in OOB is sufficient to deal with the change and maintain the predictive performance; when using OOB with other active concept drift detectors, the number of false alarms and performance resetting need to be carefully considered.

\vspace{3mm}
\noindent C.2. $\mathbf p\left(\mathbf{x} \mid y\right)$ \textbf{Concept Drift}

The data streams in this section only involve $p\left( \mathbf{x} \mid y \right)$ type of concept drift, without $P\left( y \right)$ and $P\left(  y \mid \mathbf{x} \right)$ changes. The class imbalance ratio is fixed to 1:9 and we let the positive class be the minority, so that the data stream is constantly imbalanced. The concept drift in each data stream is controlled by $p\left ( \mathbf{x} \right )$ of the negative class, as shown in Table~\ref{tab:pxydata}.

Table~\ref{tab:pxyDetectors} compares the detection performance of the three active concept drift detectors. Similar to our previous results, DDM-OCI and LFR are more sensitive to $P\left( x \mid y \right)$ changes than PAUC-PH. When DDM-OCI and LFR work with OOB, their TDR shows 100\%; and LFR has higher FA and shorter DOD than DDM-OCI, due to more indicators it monitors. PAUC-PH shows 0\% TDR in most cases of working with both OB and OOB. Different from $P\left( y \right)$ changes, when DDM-OCI and LFR work with OB, their TDR is rather low, which suggests that their sensitivity is dependent on the class imbalance techniques. Unlike the cases with class imbalance changes, where it is possible for the minority-class examples to become more frequent, the data streams generated in this section have a fixed minority class with a constantly small prior probability. In other words, it would be more difficult to recognize examples from this minority class, which indirectly affects the detection sensitivity of DDM-OCI and LFR. When oversampling is applied, which introduces more training examples for the minority class, the performance metrics (G-mean, recall and precision) monitored by DDM-OCI and LFR can be substantially improved. It also increases the possibility of reporting a concept drift. This explains the low detection rate of DDM-OCI and LFR when working with OB and their high detection rate when working with OOB. 

\begin{table}[htp]
\caption{Performance of the 3 active concept drift detectors on artificial data with $p\left(\mathbf{x} \mid y\right)$ changes: TDR, FA and DoD. The `-' symbol indicates that no concept drift is detected.}
\label{tab:pxyDetectors}
\centering
\begin{tabular}{|c|c|c|c|c|}
\hline
 & Method & TDR & FA & DoD \\ \hline
 \multirow{6}{*}{\begin{turn}{90}SINE1\end{turn}} & DDM-OCI+OB & 0\% & 0 & - \\ \cline{2-5}
&DDM-OCI+OOB & 100\% & 1.25 & 594 \\ \cline{2-5}
&LFR+OB & 0\% & 0.05 & -\\ \cline{2-5}
&LFR+OOB & 100\% & 3.99 & 528\\ \cline{2-5}
&PAUC-PH+OB & 4\% & 0.45 & 232\\ \cline{2-5}
&PAUC-PH+OOB & 0\% & 0.45 & - \\ \hline

 \multirow{6}{*}{\begin{turn}{90}SINE1g\end{turn}} & DDM-OCI+OB & 0\% & 0 & - \\ \cline{2-5}
&DDM-OCI+OOB & 100\% & 1.37 & 387 \\ \cline{2-5}
&LFR+OB & 0\% & 0 & -\\ \cline{2-5}
&LFR+OOB & 100\% & 5.45 & 258 \\ \cline{2-5}
&PAUC-PH+OB & 0\% & 1.04 & -\\ \cline{2-5}
&PAUC-PH+OOB & 0\% & 1 & - \\ \hline

\multirow{6}{*}{\begin{turn}{90}SEA\end{turn}} &DDM-OCI+OB & 16\% & 1 & 1394\\ \cline{2-5}
&DDM-OCI+OOB & 100\% & 4.03 & 473\\ \cline{2-5}
&LFR+OB & 100\% & 0.31 & 52\\ \cline{2-5}
&LFR+OOB & 100\% & 13.48 & 59\\ \cline{2-5}
&PAUC-PH+OB & 0\% & 0 & -\\ \cline{2-5}
&PAUC-PH+OOB & 0\% & 0.85 & - \\ \hline

\multirow{6}{*}{\begin{turn}{90}SEAg\end{turn}} & DDM-OCI+OB & 90\% & 0.15 & 238\\ \cline{2-5}
 & DDM-OCI+OOB & 100\% & 4.03 & 279\\ \cline{2-5}
 & LFR+OB & 29\% & 0 & 1154\\ \cline{2-5}
 & LFR+OOB & 100\% & 12.75 & 196\\ \cline{2-5}
 & PAUC-PH+OB & 0\% & 1 & -\\ \cline{2-5}
 & PAUC-PH+OOB & 0\% & 1 & - \\ \hline
\end{tabular}
\end{table}

Table~\ref{tab:pxyLearners} compares recall and G-mean of all models over the new data concept. As we expected, almost all OB models show significantly worse minority-class recall and G-mean. On SINE1 and SINE1g data, minority-class recall of OB models is as low as 0, which may hinder the detection of any concept drift. Among the OOB models, those using DDM-OCI and LFR perform significantly worse than OOB using PAUC-PH and OOB itself, and the latter two show very close performance. This is because DDM-OCI and LFR trigger concept drift with false alarms, and cause model resetting multiple times. Along with the resetting, the useful and valid information learnt in the past is forgotten at the same time. For the two passive models, RLSACP and ESOS-ELM do not perform very well compared to OOB. Generally speaking, for imbalanced data streams with $p\left( \mathbf{x} \mid y \right)$ changes, class imbalance seems to be a more important issue than concept drift, considering that the learning model without triggering any concept drift detection achieves the best performance. Besides, while the adopted class imbalance technique can improve the final prediction, it can also improve the performance of active concept drift detection methods, depending on their working mechanism. 

\begin{table}[htp]
\caption{Performance of online learners on artificial data with $p\left(\mathbf{x} \mid y\right)$ changes: means and standard deviations of average recall of each class and average G-mean over the new data concept. The significantly best values among all methods are shown in bold italics.}
\label{tab:pxyLearners}
\centering
\resizebox{\columnwidth}{!}{
\begin{tabular}{|c|c|c|c|c|}
\hline
 & Method & Class+1 Recall & Class-1 Recall & G-mean \\ \hline
 \multirow{10}{*}{\begin{turn}{90}SINE1\end{turn}} & DDM-OCI+OB & 0.000$\pm$0.000 & \textbf{1.000$\pm$0.000} & 0.000$\pm$0.000 \\ \cline{2-5}
&DDM-OCI+OOB & 0.036$\pm$0.025 & 0.997$\pm$0.002 & 0.145$\pm$0.052 \\ \cline{2-5}
&LFR+OB & 0.000$\pm$0.000 & \textbf{1.000$\pm$0.000} & 0.000$\pm$0.000\\ \cline{2-5}
&LFR+OOB & 0.061$\pm$0.036 & 0.994$\pm$0.005 & 0.200$\pm$0.066\\ \cline{2-5}
&PAUC-PH+OB & 0.000$\pm$0.000 & \textbf{1.000$\pm$0.000} & 0.000$\pm$0.000\\ \cline{2-5}
&PAUC-PH+OOB & \textbf{0.689$\pm$0.038} & 0.985$\pm$0.004 & \textbf{0.811$\pm$0.027} \\ \cline{2-5}
&RLSACP & 0.090$\pm$0.028 & 0.939$\pm$0.012 & 0.251$\pm$0.045 \\ \cline{2-5}
&ESOS-ELM & 0.058$\pm$0.122 & \textbf{1.000$\pm$0.000} & 0.113$\pm$0.208 \\ \cline{2-5}
&OB & 0.000$\pm$0.000 & \textbf{1.000$\pm$0.000} & 0.000$\pm$0.000 \\ \cline{2-5}
&OOB & \textbf{0.696$\pm$0.020} & 0.985$\pm$0.004 & \textbf{0.817$\pm$0.013}\\ \hline

 \multirow{10}{*}{\begin{turn}{90}SINE1g\end{turn}} & DDM-OCI+OB & 0.000$\pm$0.000 & \textbf{1.000$\pm$0.000} & 0.000$\pm$0.000\\ \cline{2-5}
&DDM-OCI+OOB & 0.035$\pm$0.064 & 0.993$\pm$0.006 & 0.096$\pm$0.135\\ \cline{2-5}
&LFR+OB & 0.000$\pm$0.000 & \textbf{1.000$\pm$0.000} & 0.000$\pm$0.000\\ \cline{2-5}
&LFR+OOB & 0.038$\pm$0.062 & 0.992$\pm$0.008 & 0.111$\pm$0.132\\ \cline{2-5}
&PAUC-PH+OB & 0.000$\pm$0.000 & \textbf{1.000$\pm$0.000} & 0.000$\pm$0.000\\ \cline{2-5}
&PAUC-PH+OOB & \textbf{0.801$\pm$0.032} & 0.988$\pm$0.003 & \textbf{0.884$\pm$0.019}\\ \cline{2-5}
&RLSACP & 0.072$\pm$0.049 & 0.952$\pm$0.009 & 0.173$\pm$0.102\\ \cline{2-5}
&ESOS-ELM & 0.077$\pm$0.112 & 0.991$\pm$0.035 & 0.162$\pm$0.215\\ \cline{2-5}
&OB & 0.000$\pm$0.000 & \textbf{1.000$\pm$0.000} & 0.000$\pm$0.000\\ \cline{2-5}
&OOB & \textbf{0.802$\pm$0.034} & 0.988$\pm$0.003 & \textbf{0.884$\pm$0.021} \\ \hline

\multirow{10}{*}{\begin{turn}{90}SEA\end{turn}} &DDM-OCI+OB & 0.001$\pm$0.000 & \textbf{0.999$\pm$0.000} & 0.002$\pm$0.006\\ \cline{2-5}
&DDM-OCI+OOB & 0.144$\pm$0.027 & 0.973$\pm$0.007 & 0.332$\pm$0.040\\ \cline{2-5}
&LFR+OB & 0.036$\pm$0.012 & 0.984$\pm$0.005 & 0.144$\pm$0.048\\ \cline{2-5}
&LFR+OOB & 0.085$\pm$0.039 & 0.971$\pm$0.015 & 0.243$\pm$0.069\\ \cline{2-5}
&PAUC-PH+OB & 0.130$\pm$0.027 & 0.983$\pm$0.004 & 0.341$\pm$0.042\\ \cline{2-5}
&PAUC-PH+OOB & 0.459$\pm$0.044 & 0.923$\pm$0.010 & 0.645$\pm$0.030\\ \cline{2-5}
&RLSACP & 0.000$\pm$0.001 & \textbf{0.999$\pm$0.001} & 0.001$\pm$0.006\\ \cline{2-5}
&ESOS-ELM & 0.202$\pm$0.158 & 0.967$\pm$0.071 & 0.394$\pm$0.167\\ \cline{2-5}
&OB & 0.130$\pm$0.027 & 0.983$\pm$0.004 & 0.341$\pm$0.042\\ \cline{2-5}
&OOB & \textbf{0.477$\pm$0.031} & 0.919$\pm$0.010 & \textbf{0.657$\pm$0.021} \\ \hline

\multirow{10}{*}{\begin{turn}{90}SEAg\end{turn}} & DDM-OCI+OB & 0.002$\pm$0.007 & \textbf{1.000$\pm$0.000} & 0.010$\pm$0.035\\ \cline{2-5}
&DDM-OCI+OOB & 0.100$\pm$0.040 & 0.978$\pm$0.008 & 0.257$\pm$0.066\\ \cline{2-5}
&LFR+OB & 0.101$\pm$0.027 & 0.999$\pm$0.000 & 0.269$\pm$0.058\\ \cline{2-5}
&LFR+OOB & 0.050$\pm$0.029 & 0.980$\pm$0.011 & 0.182$\pm$0.065\\ \cline{2-5}
&PAUC-PH+OB & 0.107$\pm$0.025 & 0.999$\pm$0.000 & 0.278$\pm$0.046\\ \cline{2-5}
&PAUC-PH+OOB & \textbf{0.348$\pm$0.023} & 0.939$\pm$0.017 & \textbf{0.553$\pm$0.019}\\ \cline{2-5}
&RLSACP & 0.000$\pm$0.000 & \textbf{1.000$\pm$0.000} & 0.000$\pm$0.002\\ \cline{2-5}
&ESOS-ELM & 0.183$\pm$0.137 & 0.964$\pm$0.090 & 0.368$\pm$0.161\\ \cline{2-5}
&OB & 0.106$\pm$0.021 & 0.999$\pm$0.000 & 0.279$\pm$0.040\\ \cline{2-5}
&OOB & \textbf{0.345$\pm$0.027} & 0.943$\pm$0.018 & \textbf{0.552$\pm$0.022} \\ \hline
\end{tabular}
}
\end{table}

\vspace{3mm}
\noindent C.3. $\mathbf P\left(y \mid \mathbf{x}\right)$ \textbf{Concept Drift}

\begin{table*}[htp]
\caption{Artificial data streams with $P\left( y \mid \mathbf{x} \right)$ concept drift.}
\label{tab:pyxdata}
\centering
\begin{tabular}{|c|c|c|c|c|c|c|}
\hline
ID & Data& Speed & \multicolumn{2}{|c|}{Class +1} & \multicolumn{2}{|c|}{Class -1}\\ \cline{4-7}
&&& Old concept & New concept & Old concept & New concept \\ \hline
1&SINE1 & Abrupt & \multirow{2}{*}{Points below $y=\sin \left ( x \right )$} & \multirow{2}{*}{Points above/on $y=\sin \left ( x \right )$} & \multirow{2}{*}{Points above/on $y=\sin \left ( x \right )$} & \multirow{2}{*}{Points below $y=\sin \left ( x \right )$} \\ \cline{1-3}
2&SINE1g & Gradual &&&& \\ \hline
3&SEA & Abrupt & \multirow{2}{*}{$x_1+x_2 \leq 7$} & \multirow{2}{*}{$x_1+x_2 \leq 13$} & \multirow{2}{*}{$x_1+x_2 > 7$} & \multirow{2}{*}{$x_1+x_2 > 13$} \\ \cline{1-3}
4&SEAg & Gradual &&&&\\ \hline
\end{tabular}
\end{table*}

The data streams in this section only involve $P\left( y \mid \mathbf{x} \right)$ type of concept drift, without $P\left( y \right)$ and $p\left(  \mathbf{x} \mid y \right)$ changes. Following the settings in Section~\ref{subsec:artificial_analysis}.2, we fix the class imbalance ratio to 1:9 and let the positive class be the minority, so that the data stream is constantly imbalanced. As shown in Table~\ref{tab:pyxdata}, the data distribution in SINE1 and SINE1g involves a concept swap, and this change occurs probabilistically in SINE1g; the data distribution in SEA and SEAg has a concept threshold moving, and this change occurs continuously in SEAg. The change in SEA and SEAg is less severe than the change in SINE1 and SINE1g, because some of the examples from the old concept are still valid under the new concept after the threshold moves completely. The concept drift discussed in this section belongs to the real concept drift category, which affects the classification boundary and is expected to be captured by all concept drift detectors. 

According to Table~\ref{tab:pyxDetectors}, we can see that DDM-OCI and LFR have difficulty in detecting the concept drift when working with OB, because of the poor recall and G-mean produced by OB, which is also observed and explained in Section~\ref{subsec:artificial_analysis}.2. When DDM-OCI and LFR work with OOB, their detection rate TDR is greatly improved (above 90\% in most cases). This is because the improved performance metrics facilitate the detection. LFR is more sensitive to the change, which produces higher FA and shorter DoD. Different from previous observations in terms of concept drift detection performance, PAUC-PH working with OB produces 100\% TDR and low FA on data streams SINE1 and SINE1g, but PAUC-PH does not work well with OOB on the same data. It is interesting to see that oversampling does not always play a positive role in drift detection. One possible reason is that class imbalance techniques may sometimes hide the performance drop caused by the real concept drift, while it tries to maintain the overall predictive performance, especially for AUC type of metrics in our case. On data streams SEA and SEAg, PAUC-PH does not report any concept drift, probably due to the less severe concept drift. 

\begin{table}[htp]
\caption{Performance of the 3 active concept drift detectors on artificial data with $P\left(y \mid \mathbf{x}\right)$ changes: TDR, FA and DoD. The `-’ symbol indicates that no concept drift is detected.}
\label{tab:pyxDetectors}
\centering
\begin{tabular}{|c|c|c|c|c|}
\hline
 & Method & TDR & FA & DoD \\ \hline
 \multirow{6}{*}{\begin{turn}{90}SINE1\end{turn}} & DDM-OCI+OB & 0\% & 0 & - \\ \cline{2-5}
&DDM-OCI+OOB & 97\% & 1.02 & 1166 \\ \cline{2-5}
&LFR+OB & 0\% & 0 & -\\ \cline{2-5}
&LFR+OOB & 91\% & 3.92 & 783\\ \cline{2-5}
&PAUC-PH+OB & 100\% & 1.03 & 884\\ \cline{2-5}
&PAUC-PH+OOB & 2\% & 1.28 & 1180 \\ \hline

 \multirow{6}{*}{\begin{turn}{90}SINE1g\end{turn}} & DDM-OCI+OB & 0\% & 0 & - \\ \cline{2-5}
&DDM-OCI+OOB & 69\% & 2.16 & 165 \\ \cline{2-5}
&LFR+OB & 0\% & 1 & -\\ \cline{2-5}
&LFR+OOB & 85\% & 6.21 & 306 \\ \cline{2-5}
&PAUC-PH+OB & 100\% & 1.03 & 1119\\ \cline{2-5}
&PAUC-PH+OOB & 0\% & 1 & - \\ \hline

\multirow{6}{*}{\begin{turn}{90}SEA\end{turn}} &DDM-OCI+OB & 61\% & 0.39 & 23\\ \cline{2-5}
&DDM-OCI+OOB & 100\% & 3.87 & 151\\ \cline{2-5}
&LFR+OB & 10\% & 0.02 & 865\\ \cline{2-5}
&LFR+OOB & 100\% & 13.73 & 65\\ \cline{2-5}
&PAUC-PH+OB & 0\% & 1 & -\\ \cline{2-5}
&PAUC-PH+OOB & 0\% & 1 & -\\ \hline

\multirow{6}{*}{\begin{turn}{90}SEAg\end{turn}} & DDM-OCI+OB & 100\% & 0 & 71\\ \cline{2-5}
 & DDM-OCI+OOB & 100\% & 3.9 & 342\\ \cline{2-5}
 & LFR+OB & 3\% & 0.02 & 1036\\ \cline{2-5}
 & LFR+OOB & 100\% & 13.59 & 123\\ \cline{2-5}
 & PAUC-PH+OB & 0\% & 1 & -\\ \cline{2-5}
 & PAUC-PH+OOB & 0\% & 1 & - \\ \hline
\end{tabular}
\end{table}

\begin{table}[htp]
\caption{Performance of online learners on artificial data with $P\left(y \mid \mathbf{x}\right)$ changes: means and standard deviations of average recall of each class and average G-mean over the new data concept. The significantly best values among all methods are shown in bold italics.}
\label{tab:pyxLearners}
\centering
\resizebox{\columnwidth}{!}{
\begin{tabular}{|c|c|c|c|c|}
\hline
 & Method & Class+1 Recall & Class-1 Recall & G-mean \\ \hline
 \multirow{10}{*}{\begin{turn}{90}SINE1\end{turn}} & DDM-OCI+OB & 0.000$\pm$0.000 & \textbf{1.000$\pm$0.000} & 0.000$\pm$0.000 \\ \cline{2-5}
&DDM-OCI+OOB & 0.004$\pm$0.003 & 0.998$\pm$0.002 & 0.030$\pm$0.016 \\ \cline{2-5}
&LFR+OB & 0.000$\pm$0.000 & \textbf{1.000$\pm$0.000} & 0.000$\pm$0.000\\ \cline{2-5}
&LFR+OOB & 0.013$\pm$0.010 & 0.996$\pm$0.006 & 0.062$\pm$0.036\\ \cline{2-5}
&PAUC-PH+OB & 0.000$\pm$0.000 & \textbf{1.000$\pm$0.000} & 0.000$\pm$0.000\\ \cline{2-5}
&PAUC-PH+OOB & \textbf{0.031$\pm$0.013} & 0.941$\pm$0.009 & \textbf{0.098$\pm$0.026} \\ \cline{2-5}
&RLSACP & 0.000$\pm$0.001 & \textbf{0.999$\pm$0.001} & 0.003$\pm$0.010 \\ \cline{2-5}
&ESOS-ELM & 0.000$\pm$0.000 & 0.997$\pm$0.003 & 0.000$\pm$0.000 \\ \cline{2-5}
&OB & 0.000$\pm$0.000 & \textbf{1.000$\pm$0.000} & 0.000$\pm$0.000 \\ \cline{2-5}
&OOB & \textbf{0.033$\pm$0.012} & 0.942$\pm$0.009 & \textbf{0.102$\pm$0.022}\\ \hline

 \multirow{10}{*}{\begin{turn}{90}SINE1g\end{turn}} & DDM-OCI+OB & 0.000$\pm$0.000 & \textbf{1.000$\pm$0.000} & 0.000$\pm$0.000\\ \cline{2-5}
&DDM-OCI+OOB & 0.014$\pm$0.017 & 0.993$\pm$0.006 & 0.069$\pm$0.074\\ \cline{2-5}
&LFR+OB & 0.000$\pm$0.000 & \textbf{1.000$\pm$0.000} & 0.000$\pm$0.000\\ \cline{2-5}
&LFR+OOB & 0.019$\pm$0.018 & 0.993$\pm$0.006 & 0.086$\pm$0.077\\ \cline{2-5}
&PAUC-PH+OB & 0.000$\pm$0.000 & \textbf{1.000$\pm$0.000} & 0.000$\pm$0.000\\ \cline{2-5}
&PAUC-PH+OOB & \textbf{0.031$\pm$0.011} & 0.993$\pm$0.002 & \textbf{0.103$\pm$0.026}\\ \cline{2-5}
&RLSACP & 0.000$\pm$0.001 & \textbf{1.000$\pm$0.000} & 0.001$\pm$0.008\\ \cline{2-5}
&ESOS-ELM & 0.000$\pm$0.000 & 0.907$\pm$0.140 & 0.000$\pm$0.000\\ \cline{2-5}
&OB & 0.000$\pm$0.000 & \textbf{1.000$\pm$0.000} & 0.000$\pm$0.000\\ \cline{2-5}
&OOB & 0.027$\pm$0.010 & 0.995$\pm$0.002 & 0.093$\pm$0.028 \\ \hline

\multirow{10}{*}{\begin{turn}{90}SEA\end{turn}} &DDM-OCI+OB & 0.013$\pm$0.022 & \textbf{0.999$\pm$0.001} & 0.050$\pm$0.085\\ \cline{2-5}
&DDM-OCI+OOB & 0.110$\pm$0.031 & 0.968$\pm$0.008 & 0.311$\pm$0.057\\ \cline{2-5}
&LFR+OB & 0.149$\pm$0.025 & \textbf{0.999$\pm$0.000} & 0.378$\pm$0.036\\ \cline{2-5}
&LFR+OOB & 0.031$\pm$0.022 & 0.964$\pm$0.016 & 0.144$\pm$0.071\\ \cline{2-5}
&PAUC-PH+OB & 0.153$\pm$0.023 & \textbf{0.999$\pm$0.000} & 0.384$\pm$0.031\\ \cline{2-5}
&PAUC-PH+OOB & \textbf{0.292$\pm$0.017} & 0.967$\pm$0.008 & \textbf{0.530$\pm$0.015}\\ \cline{2-5}
&RLSACP & 0.013$\pm$0.013 & 0.995$\pm$0.001 & 0.072$\pm$0.063\\ \cline{2-5}
&ESOS-ELM & 0.065$\pm$0.068 & 0.997$\pm$0.022 & 0.222$\pm$0.106\\ \cline{2-5}
&OB & 0.152$\pm$0.023 & \textbf{0.999$\pm$0.000} & 0.383$\pm$0.032\\ \cline{2-5}
&OOB & 0.287$\pm$0.014 & 0.966$\pm$0.008 & 0.525$\pm$0.012 \\ \hline

\multirow{10}{*}{\begin{turn}{90}SEAg\end{turn}} & DDM-OCI+OB & 0.000$\pm$0.002 & \textbf{1.000$\pm$0.000} & 0.001$\pm$0.013\\ \cline{2-5}
&DDM-OCI+OOB & 0.042$\pm$0.022 & 0.988$\pm$0.006 & 0.163$\pm$0.059\\ \cline{2-5}
&LFR+OB & 0.145$\pm$0.032 & 0.999$\pm$0.000 & 0.356$\pm$0.066\\ \cline{2-5}
&LFR+OOB & 0.024$\pm$0.018 & 0.985$\pm$0.006 & 0.112$\pm$0.065\\ \cline{2-5}
&PAUC-PH+OB & 0.152$\pm$0.019 & 0.999$\pm$0.000 & 0.370$\pm$0.027\\ \cline{2-5}
&PAUC-PH+OOB & \textbf{0.288$\pm$0.034} & 0.974$\pm$0.010 & \textbf{0.512$\pm$0.036}\\ \cline{2-5}
&RLSACP & 0.009$\pm$0.018 & \textbf{1.000$\pm$0.000} & 0.043$\pm$0.077\\ \cline{2-5}
&ESOS-ELM & 0.138$\pm$0.088 & 0.993$\pm$0.057 & 0.336$\pm$0.106\\ \cline{2-5}
&OB & 0.149$\pm$0.025 & 0.999$\pm$0.000 & 0.364$\pm$0.042\\ \cline{2-5}
&OOB & \textbf{0.282$\pm$0.032} & 0.974$\pm$0.008 & \textbf{0.506$\pm$0.034} \\ \hline
\end{tabular}
}
\end{table}

\begin{figure*}[htp]
\centerline{
\includegraphics[width=2.5in]{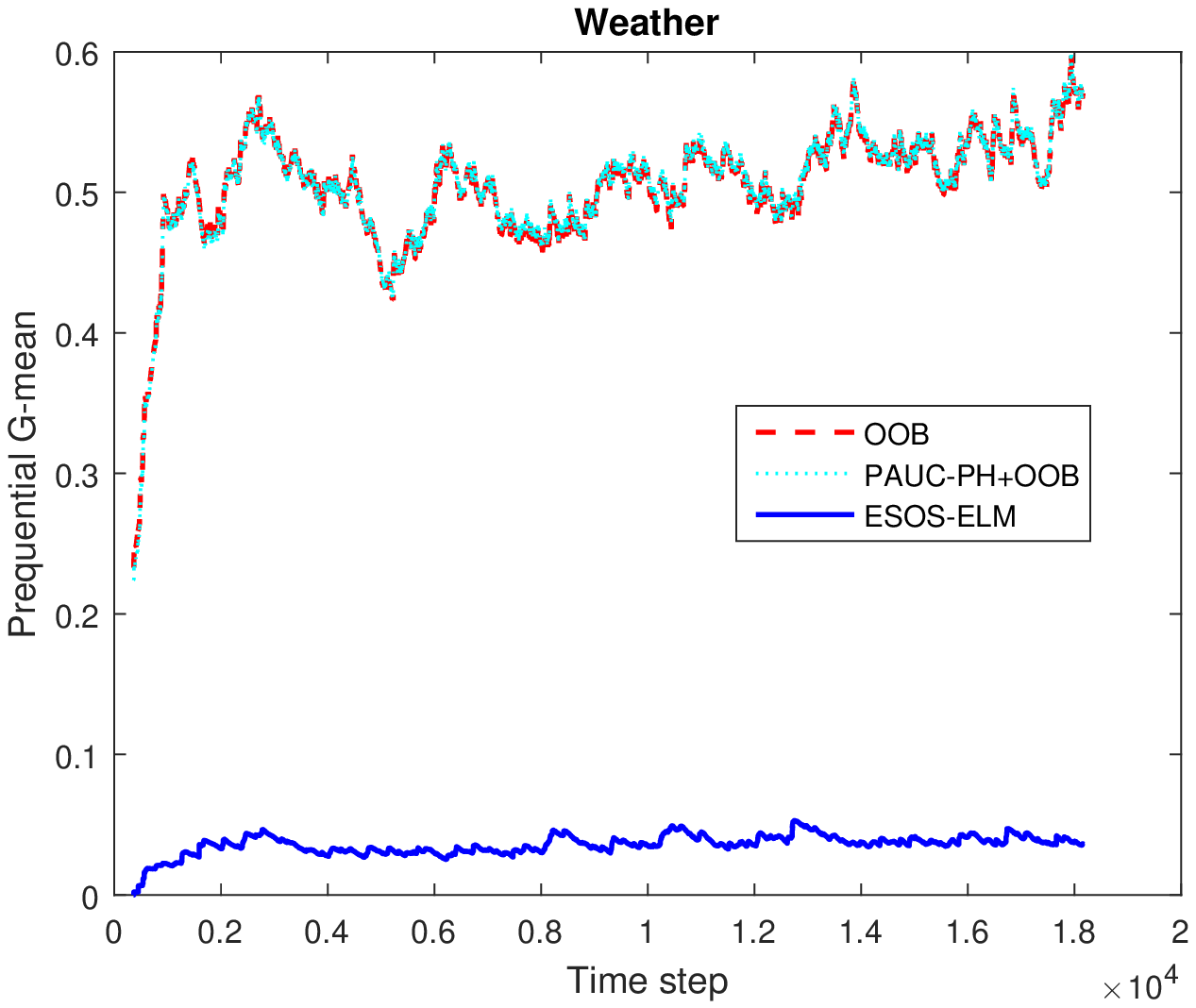}
\includegraphics[width=2.5in]{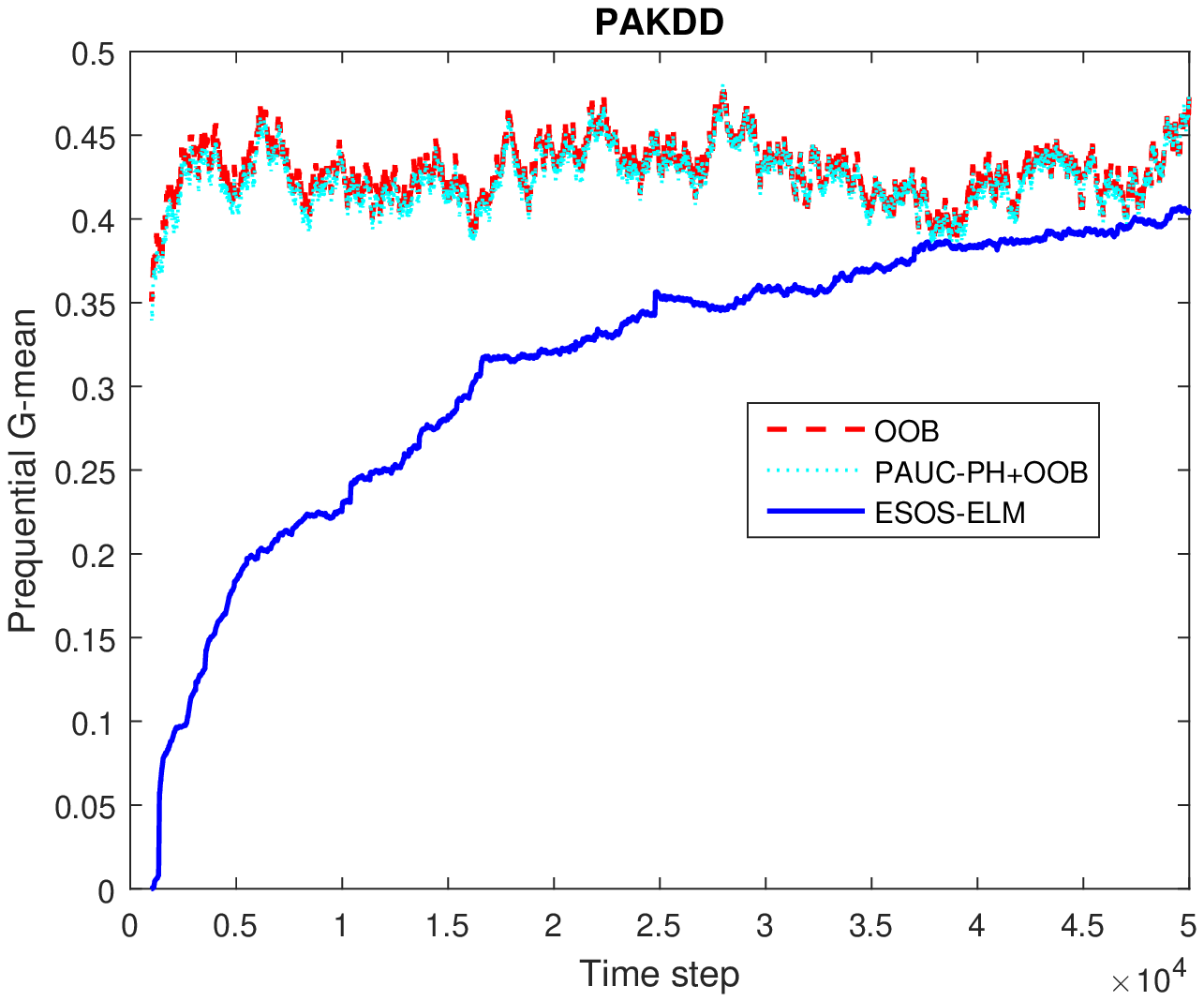}
\includegraphics[width=2.5in]{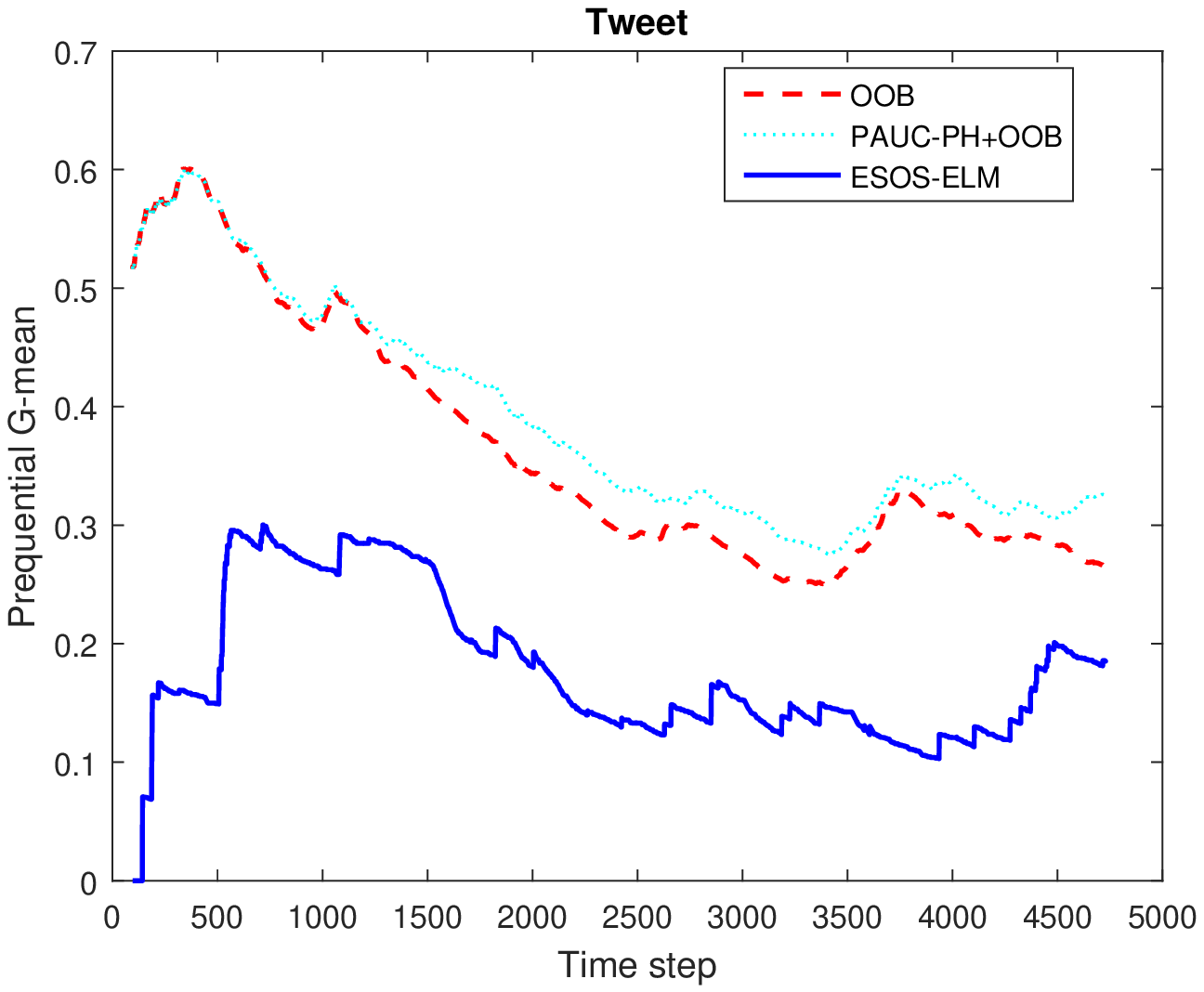}}
\caption{Time-decayed G-mean curves (decay factor = 0.995) from OOB, PAUC-PH+OOB and ESOS-ELM on real-world data.}
\label{fig:realdata}
\end{figure*}

The recall and G-mean over the new data concept in Table~\ref{tab:pyxLearners} further confirms the above analysis. The OB models produce very low minority-class recall and thus low G-mean. RLSACP and ESOS-ELM do not perform well on the new data concept either. By comparing the models that captures concept drifts (DDM-OCI+OOB, LFR+OOB, PAUC-PH+OB) and the models without reporting any concept drift (PAUC-PH+OOB, OOB), it seems that class imbalance causes a more difficult learning issue than the real concept drift in our cases. The models solely tackling class imbalance produce the significantly best recall and G-mean. The rather low imbalance ratio (i.e. 1:9) could be a reason. It would be worth discussing various imbalance levels in data with concept drift in our future work, in order to find out when it is worthwhile considering concept drift in imbalanced data streams. By comparing the results in Table~\ref{tab:pyxLearners}, Table~\ref{tab:pxyLearners} and Table~\ref{tab:pyLearners}, the $P\left( y \mid \mathbf{x} \right)$ type of concept drift indeed leads to the most performance reduction. It is consistent with our understanding that the real concept drift is the most radical type of change in data. However, existing approaches do not seem to tackle it well when data streams are very imbalanced. To develop better concept drift detection methods, the key issues here include how to best have them and class imbalance techniques work together and how to tackle the performance loss brought by false alarms. 

\subsection{Comparative Study on Real-World Data}
\label{subsec:real_analysis}

After the detailed analysis of the three types of concept drift, we now look into the performance of the above learning models on the three real-world data sets (PAKDD~\cite{Linhart2010ln}, Weather~\cite{Ditzler2013nh} and Tweet~\cite{LiWDWC12}) described in Section~\ref{subsec:data}. Based on the experimental results on the artificial data, we focus on the best active (PAUC+OOB) and the best passive concept drift detection methods (ESOS-ELM) here for a clear observation, in comparison with OOB. The three methods use the same parameter settings as before. The initialisation and validation data required by ESOS-ELM is the first 2\% examples of each data set. 

Without knowing the true concept drifts in real-world data, we calculate and track the time-decayed G-mean by setting the decay factor to 0.995, which means that the old performance is forgotten at the rate of 0.5\%. All the compared metrics are the average of 100 runs in the following figures.

Fig.~\ref{fig:realdata} presents the time-decayed G-mean curves from OOB, PAUC-PH+OOB and ESOS-ELM on the three real-world data sets. The average number of reported drift by PAUC-PH is 1, 3 and 1 on Weather, PAKDD and Tweet data respectively. Compared to the artificial cases, we obtain some similar results: the passive approach ESOS-ELM does not perform as well as the other two methods; OOB and PAUC-PH show very close G-mean over time on Weather and PAKDD data, which suggests the importance of tackling class imbalance adaptively. 

In the PAKDD plot, we can see that the G-mean level is relatively stable without significant drop; differently, G-mean in the Tweet plot is reducing. It may suggest that the concept drift in PAKDD is less significant or influential than that in Tweet. Compared to the gradual market and environment change in PAKDD, the tweet topic change can be much faster and more noticeable. Therefore, although PAUC-PH detects 3 concept drifts in PAKDD, the two methods, OOB and PAUC-PH+OOB, does not show much difference. In tweet, PAUC-PH+OOB presents better G-mean than using OOB alone, showing the positive effect of the active concept drift detector in fast changing data streams. 

\subsection{Further Discussions}
\label{subsec:discussion}

In this section, we summarize and further discuss the results in the above comparative study on the artificial and real-world data. We also answer the research questions proposed at the beginning of this paper. When dealing with imbalanced data streams with concept drift, we have obtained the following:

\begin{itemize}
\item When both class imbalance and concept drift exist, class imbalance status and class imbalance changes are shown to be more crucial issues than the traditional concept drift (i.e. $p\left( \mathbf{x} \mid y \right)$ and $P\left( y \mid \mathbf{x} \right)$ changes) in terms of the online prediction performance. It is necessary to adopt adaptive class imbalance techniques (e.g. OOB discussed in our experiment), in addition to using concept drift detection methods alone (e.g. DDM-OCI, LFR). Most existing papers that proposed new concept drift detection methods for imbalanced data so far did not consider the effect of class imbalance techniques on final prediction and concept drift detection. 
\item $P\left( y \mid \mathbf{x} \right)$ concept drift (i.e. real concept drift) is the most severe type of change in data, compared to  $p\left( \mathbf{x} \mid y \right)$ and $P\left( y \right)$ concept drift. This is based on the observation on the final prediction performance. For all three types of concept drift, existing concept drift approaches do not show much benefit in performance improvement. Concept drift is hard to be detected when no class imbalance technique is applied. Their drift detection performance is affected by the class imbalance technique, depending on their detection mechanism. 
\item For $P\left( y \right)$ concept drift, it is not necessary to apply any concept drift detection methods that are not designed for class imbalance changes, due to their false alarms and model resetting. It is crucial to detect and handle the class imbalance change in time. 
\item From the results on real-world data, we see that the effectiveness of traditional concept drift detectors (e.g. PAUC-PH) depends on the type of concept drift. For fast and significant concept drift, applying PAUC-PH seems to be more beneficial to the prediction performance.
\item Among existing methods designed for imbalanced data with concept drift (4 active methods and 2 passive methods), the passive methods (i.e. ESOS-ELM and RLSACP) do not perform well in general. Although they contain both class imbalance and concept drift techniques, firstly, their class imbalance technique is not effectively adaptive to class imbalance changes, so that wrong imbalance status might be used during learning; secondly, they are restricted to the use of certain perceptron-based classifiers, so that the disadvantages of the classifiers are also inherited by the online model. For example, the training of OS-ELM in ESOS-ELM requires initialisation and validation data sets reflecting the correct data concepts, and the weighted OS-ELM was found to over-emphasize the minority class and present large performance variance sometimes in earlier studies~\cite{Wang2014vb}. 
\item Among the three active methods discussed in this work, which are DDM-OCI, LFR and PAUC-PH, DDM-OCI and LFR are more sensitive to concept drift than PAUC-PH, with a higher detection rate but also higher false alarms. In addition, the detection performance of DDM-OCI and LFR can be greatly improved by OOB. The explanation can be found in the previous analysis. 
\end{itemize}

Overall, all these results suggest us that class imbalance and concept drift need to be studied simultaneously, when we design an algorithm to deal with imbalanced data with concept drift. Their mutual effect must be taken into consideration. Hence, we propose the following key issues to be considered for an effective algorithm:

\begin{itemize}
\item Is the class imbalance technique effective in predicting minority-class examples?
\item Is the class imbalance technique adaptive to class imbalance changes?
\item Is the concept drift technique effective in detecting different types of concept drift, in terms of detection rate, false alarms and detection promptness? Which type of concept drift is it designed for? Which type of concept drift does it perform better?
\item Is the detection performance of the concept drift technique affected by the class imbalance technique? And how?
\item How can we have the class imbalance technique and concept drift technique work together, to achieve better detection rate, fewer false alarms, less detection delay or better online prediction?
\end{itemize}

\section{Conclusion}
\label{sec:con}
This paper gives the first systematic study of handling concept drift in class-imbalanced data streams. In the context of online learning, we provide a thorough review and an experimental insight into this problem. 

First, a comprehensive review is given, including the problem description and definitions, the individual learning issues and solutions in class imbalance and concept drift respectively, the combined challenges and existing solutions in online class imbalance learning with concept drift, and example applications. The review reveals research gaps in the field of online class imbalance learning with concept drift. Specifically, little work has looked into the concept drift issue in imbalanced data streams systematically, although a few methods have been proposed for this purpose; $P\left( y \right)$ type of concept drift is closely related to the class imbalance issue, but it has not been investigated properly so far; most existing concept drift detection methods are only designed for or tested on balanced data streams. 

Second, to fill in these research gaps, we carry out a thorough empirical study by looking into the following research questions: 1) what are the challenges in detecting each type of concept drift when the data stream is imbalanced (i.e. changes in $P\left( y \right)$, $p\left( \mathbf{x} \mid y \right)$, and $P\left( y \mid \mathbf{x} \right)$)? 2) Among the proposed methods designed for online class imbalance learning with concept drift, i.e. DDM-OCI~\cite{Wang2013bp}, LFR~\cite{Wang2015zo}, PAUC-PH~\cite{Brzezinski2015ib}, OOB~\cite{Wang2014vb}, RLSACP~\cite{Ghazikhani2013wm} and ESOS-ELM~\cite{Mirza2015nt}, which one performs better for which type of concept drift? 3) Would applying class imbalance techniques (e.g. resampling methods) facilitate the concept drift detection and online prediction? By generating artificial data streams with different types of class imbalance and concept drift and experimenting on real-world data, we make the following conclusions.

For the first research question, a $P\left( y \right)$ change can be easily tackled by an adaptive class imbalance technique (e.g. OOB used in this work). The traditional concept drift detectors, such as LFR, DDM-OCI and PAUC-PH, do not perform well in detecting a $p\left( \mathbf{x} \mid y \right)$ change. The prediction performance on an imbalanced data stream with $p\left( \mathbf{x} \mid y \right)$ changes can be effectively improved by solely using an adaptive class imbalance technique. A $P\left( y \mid \mathbf{x} \right)$ change is the most challenging case for learning, where the traditional active and passive concept drift detection methods do not bring much performance improvement. Class imbalance is shown to be a more crucial issue in terms of final prediction performance. 

For the second research question, the two passive methods, RLSACP and ESOS-ELM, do not perform well in general. DDM-OCI and LFR are sensitive to different types of concept drift, with a high detection rate but also high false alarms. PAUC-PH is more conservative in terms of drift detection. Based on the observation on minority-class recall and G-mean, the combination PAUC-PH and OOB was shown to be the best approach among all.

For the third research question, it is necessary to apply adaptive class imbalance techniques when learning from imbalanced data streams with concept drift -- they bring the most prediction performance improvement. In our experiment, our class imbalance technique OOB facilitates the concept drift detection of DDM-OCI and LFR. 

This paper also provides guidelines for future algorithm design. Several important issues are pointed out for consideration. There are still many challenges and learning issues in this field that are worth of ongoing research, such as more effective concept drift detection methods for imbalanced data streams, studying the mutual effect of class imbalance and concept drift, and more real-world applications with different types of class imbalance and concept drift.



\bibliographystyle{IEEEtran}
\bibliography{reference}

%








\end{document}